\documentclass{article}
\usepackage{iclr2027_conference,times}

\usepackage{amsmath,amssymb,bm}
\usepackage{booktabs}
\usepackage{multirow}
\usepackage{algorithm}
\usepackage{algpseudocode}
\usepackage{float}
\usepackage{graphicx}
\graphicspath{{./}}
\usepackage[hyphens]{url}
\usepackage{hyperref}
\usepackage{caption}
\usepackage[table]{xcolor}
\usepackage{xspace}

\usepackage{amsmath,amsfonts,bm}

\def\eqref#1{equation~\ref{#1}}

\def\1{\bm{1}}

\DeclareMathAlphabet{\mathsfit}{\encodingdefault}{\sfdefault}{m}{sl}
\SetMathAlphabet{\mathsfit}{bold}{\encodingdefault}{\sfdefault}{bx}{n}

\newcommand{\car}{RFPO\xspace}
\newcommand{\vB}{\bar{V}_\phi}
\newcommand{\rhat}{\hat{r}}
\newcommand{\rstar}{r^{\star}}

\definecolor{hdrbg}{gray}{0.92}
\definecolor{carbg}{RGB}{232,240,250}

\title{Unlocking the Critic:\\Reward-Free Policy Optimization for LLM Post-Training}
\author{Hongyang (Kevin) Li$^{1}$\quad Xiao Li$^{2}$\quad Caesar Wu$^{1}$\quad Said Mammar$^{3}$\quad Gr\'egoire Danoy$^{1}$\quad Pascal Bouvry$^{1}$}
\newcommand{\affilnote}{\begingroup\renewcommand{\thefootnote}{}\footnotetext{\raggedright
$^{1}$University of Luxembourg\par
\hspace*{0.6em}\texttt{\{hongyang.li,\,caesar.wu,\,gregoire.danoy,\,pascal.bouvry\}@uni.lu}\par
$^{2}$Seafill Open-Source Community, \texttt{xiao.li@seafill.com}\par
$^{3}$Universit\'e Paris-Saclay, \texttt{said.mammar@univ-evry.fr}}\endgroup}
\iclrfinalcopy

\begin{document}
\raggedbottom
\maketitle
\affilnote

\begin{abstract}
Recent approaches to reinforcement learning (RL) post-training for large language models increasingly remove the critic to reduce training instability and memory overhead. Even where a critic is trained, it is discarded once training ends, although it has learned to predict outcomes. We revisit this trend and show that a pretrained critic's ability to predict future outcomes can make it a valuable asset for efficient long-horizon reasoning. First, we find that instability in critic-based RL for long chain-of-thought reasoning is largely an optimization artifact: keeping policy updates small and low in variance restores stable convergence. Second, a well-pretrained critic estimates the posterior probability of eventual success from later trajectory states and unfinished prefixes. Its predictions therefore provide outcome-derived, dense, per-prefix learning signals that, during policy optimization, require neither completed rollouts, step-level annotations, nor external reward labels. Building on this insight, we introduce Reward-Free Policy Optimization (RFPO), which repurposes a single calibrated, frozen critic as a rollout-level reward, a value baseline for generalized advantage estimation, and a success forecaster for unfinished prefixes. We further show that binarizing the debiased score stops the policy from exploiting the critic's length bias. Binarized, RFPO matches supervised PPO without a single label in the training loop, while substantially cutting compute and memory overhead. This makes RFPO particularly well suited to long-horizon reasoning tasks, where outcomes arrive late and generation dominates cost: because rollouts can be rewarded before they finish, training no longer has to pay for waiting on every trajectory to complete. Our findings challenge the prevailing critic-free paradigm and establish critic-based, reward-free optimization as a scalable and computationally efficient path for LLM post-training.

\end{abstract}

\section{Introduction}
\label{sec:intro}

\looseness=-1
Reinforcement learning is now the dominant recipe for turning a pretrained
language model into a reasoner
\citep{deepseek2025r1,grattafiori2024llama3,lambert2024tulu}, and the recipe has
converged on a particular shape: PPO's value network \citep{schulman2017ppo}---a
second model the size of the policy---is replaced by a baseline computed from
several samples of the same prompt
\citep{shao2024deepseekmath,ahmadian2024back,yu2025dapo,hu2025reinforcepp,liu2025drgrpo,zheng2025gspo}.
Two arguments carried this change: the critic is expensive to hold and to train,
and its estimates were judged too unreliable to be worth the trouble
\citep{kazemnejad2024vineppo}. Recent reasoning and agentic systems are trained
with no value network at all
\citep{deepseek2025r1,kimi2025k15,yang2025qwen3,kimi2025k2,minimax2025m1}.
A counter-current has begun to repair the critic instead of removing it, by
pretraining the value head, by making it generative, or by weighting it by how
much of the return it explains
\citep{yuan2025vcppo,yue2025vapo,seed2025thinking,shan2026genac,pan2026evpo}---but every one of
these works keeps the critic in the job it already had, reducing the variance of
the policy gradient, while the reward still comes from outside. We ask a
different question. If a critic can be trained well enough to be trusted as a
baseline, is it also good enough to be the \emph{reward}? The question matters because the reward is where these pipelines run out of
resources. Human feedback, outcome and process reward models each require their
own annotation project
\citep{christiano2017deep,ouyang2022training,cobbe2021training,uesato2022solving,lightman2024lets,wang2024mathshepherd};
verifiable rewards avoid the reward model but require a reference answer for
every training prompt \citep{yu2025dapo,lambert2024tulu}. Each route buys the
same thing---a number attached to every rollout---at a price that does not scale
with the amount of RL one would like to run.

\begin{figure}[t]
\centering
\includegraphics[width=\textwidth]{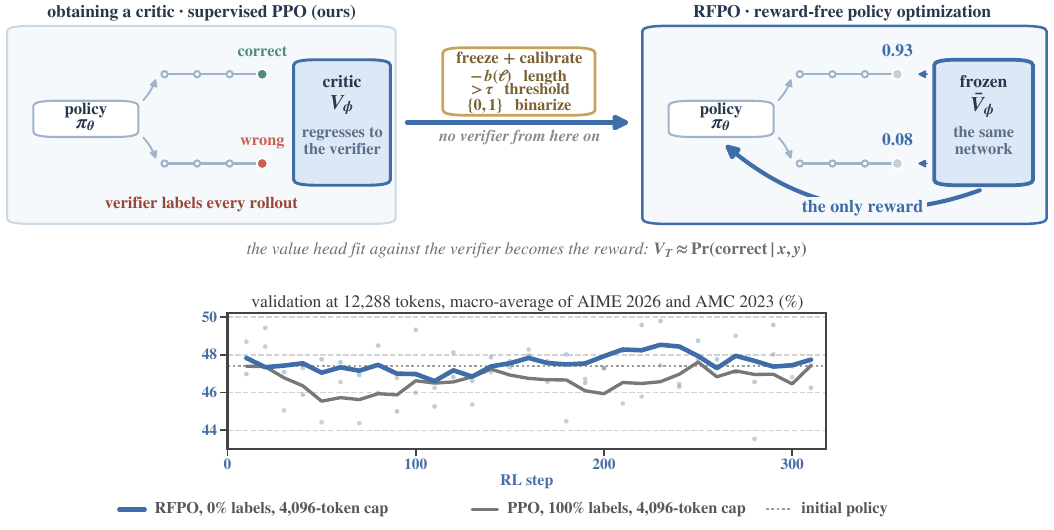}
\caption{\looseness=-1 \textbf{Overview of RFPO.} Top left: a critic is obtained by fitting
a value network against a verifier, here within a supervised PPO run. Top right:
frozen and calibrated once, it becomes the only reward and the GAE baseline for
policy optimization. Bottom: validation at $12{,}288$ tokens (macro-average of AIME 2026 and
AMC 2023) of \car and supervised PPO, both trained under a $4{,}096$-token cap
that leaves about half of \car's training rollouts unfinished; each dot is one validation
evaluation (every 10 steps, 8 samples per problem) and each line a centered
moving average over five consecutive evaluations, i.e.\ $\pm 20$ steps
(\S\ref{sec:results}).}
\label{fig:overview}
\end{figure}

\looseness=-1
Under the sparse terminal
correctness rewards used in reasoning RL, with the standard
$\gamma{=}\lambda{=}1$, the critic's regression target at every prefix is the
trajectory's final reward, so its fixed point is $V_\phi(x,y_{\le t}) \approx
\Pr(\text{success}\mid x, y_{\le t})$ \citep{sutton2018reinforcement}. A well-pretrained critic approximates this closely enough over the later part
of a rollout, not only at its final token, for its value to serve as the
reward. From an ordinary supervised
PPO run on mathematical reasoning we take an actor and a critic checkpoint,
freeze the critic, calibrate its value, and continue policy optimization with
that calibrated value as the only reward. The same frozen network also serves as
the GAE baseline \citep{schulman2016gae} and, at unfinished prefixes, as a
forecast of success, so the continued stage trains no value network and waits on
no verifier (Figure~\ref{fig:overview}).

\looseness=-1
Our contributions: \textbf{(i)~Critic instability is an optimization artifact.} The instability
that pushed critics out of long chain-of-thought RL
\citep{yuan2025vcppo,yue2025vapo} comes from the update recipe, not the value
network: it recedes once each policy update is kept small and low in variance
(\S\ref{sec:setup}, App.~\ref{app:inner}).
\textbf{(ii)~One frozen critic, three roles.} A well-pretrained critic estimates
the posterior probability of success from later states and unfinished prefixes.
RFPO uses that single network as the rollout-level reward, the GAE baseline, and
a forecaster for unfinished prefixes, giving dense, per-prefix signals without
costly external labels during policy optimization (\S\ref{sec:method}).
\textbf{(iii)~Calibration guards against length bias.} The critic's score
carries a length bias that a continuous reward lets the policy exploit in either
direction; binarizing the debiased score closes that channel. Used as a
continuous reward, the same critic climbs above fully supervised PPO and then
gives the gain back (\S\ref{sec:results}, App.~\ref{app:reward}).
\textbf{(iv)~Parity at lower cost, built for long horizons.} Binarized, RFPO
matches supervised PPO at a $5{,}120$-token cap with no label in the training loop while substantially
cutting compute and memory; because it scores truncated rollouts with accuracy
comparable to complete ones, it suits long-horizon reasoning, where waiting for
every trajectory to finish is what costs the most. Under a $4{,}096$-token cap
that leaves about half of every batch unfinished, it validates above supervised
PPO trained under the same cap on 19\% fewer GPU-hours, without verifier labels during RL
(\S\ref{sec:results}, \S\ref{sec:trust}, Figs.~\ref{fig:overview} and~\ref{fig:cap4096}).

\section{Related Work}
\label{sec:related}

\looseness=-1
Process reward models give dense, per-step signals, but they judge the steps
already written and require step-level annotation
\citep{lightman2024lets,wang2024mathshepherd}; a critic instead forecasts where
a prefix will end and is learned from outcomes alone. Label-free alternatives
replace labels with majority voting \citep{zuo2025ttrl,wang2023selfconsistency},
self-rewarding \citep{yuan2024selfrewarding}, model confidence
\citep{zhao2025learning} or generated rubrics \citep{sheng2026rlcer}, and
without grounded supervision they are prone to reward hacking
\citep{amodei2016concrete,shao2025spurious}. Implicit-reward methods such as DPO
\citep{rafailov2023direct} and PRIME \citep{cui2025process} still need labels
during training. Our reward is grounded in verifier outcomes, obtained before
policy optimization, frozen, and bounded by calibration. As reasoning traces and agentic episodes lengthen, generation rather than
optimization dominates cost \citep{zhou2026sparrow,kimi2025k2}, and sparse
terminal rewards make credit assignment harder \citep{kazemnejad2024vineppo}
because the verifier says nothing until an answer appears. Existing remedies
truncate rollouts or sparsify attention but have no way to score an unfinished
trajectory; the critic's forecast supplies exactly that.

\section{Reward-Free Policy Optimization}
\label{sec:method}

\looseness=-1
\car optimizes a policy using a pretrained critic as its reward. Given a value
network fit to terminal correctness under $\gamma{=}\lambda{=}1$, \car freezes
it, calibrates its output once against a small sample of rollouts, and optimizes
the policy against it, so the loop contains no verifier and trains no value
network. Any procedure that yields a critic able to rank trajectories well will
do.
\S\ref{sec:identity} shows what makes a critic reliable and how ours was trained;
\S\ref{sec:calib} how it is frozen and calibrated.

\subsection{A critic worth consulting}
\label{sec:identity}\label{sec:provenance}\label{sec:trust}

\begin{figure}[t]
\centering
\includegraphics[width=\textwidth]{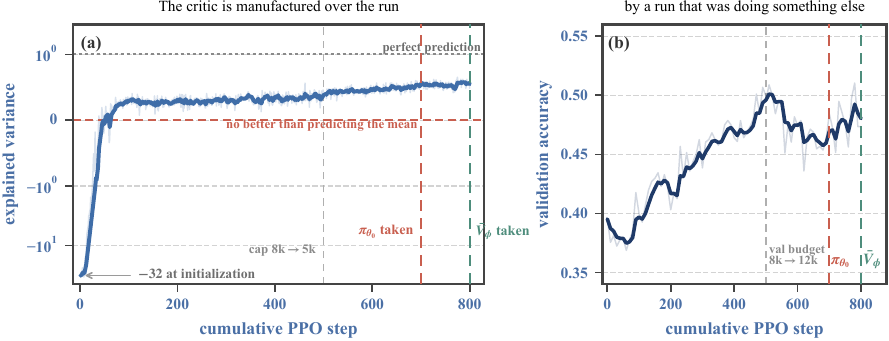}
\caption{\textbf{Training the critic.} (a) Explained variance of the value head
across both critic-pretraining phases on one cumulative axis (symmetric log
scale), rising from $-32$ at initialization to about 0.55. Across nine
checkpoints it tracks offline within-problem AUC ($r{=}0.91$; Fig.~\ref{fig:trust}d),
which makes it a cheap training-time indicator of critic quality. (b) Held-out
accuracy of the same supervised PPO run. At step 500 the training cap drops from
$8{,}192$ to $5{,}120$ tokens and the validation budget rises from $8{,}192$ to
$12{,}288$, so accuracy on either side of that line is not directly comparable.
Dashed markers show where we extract the initial policy and the critic we
freeze.}
\label{fig:provenance}
\end{figure}

\begin{figure}[t]
\centering
\includegraphics[width=\textwidth]{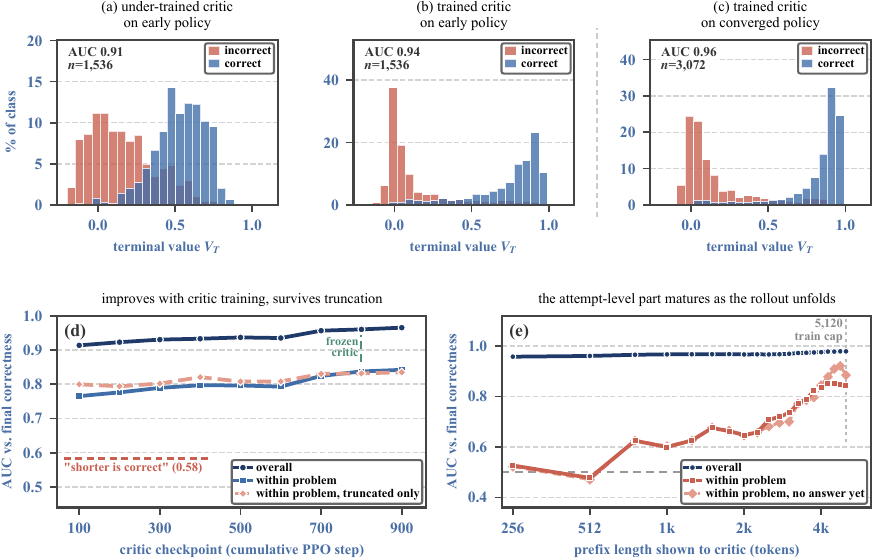}
\caption{\looseness=-1 \textbf{What the frozen critic knows.} (a--c) Terminal values of
rollouts, split by final correctness. Rollouts in (a) and (b) come from an early policy (cumulative PPO steps
100--200), those in (c) from the converged policy (steps 700--800). The critic
checkpoint at cumulative step 200 already separates correct from incorrect
early-policy rollouts,
but its values bunch between 0 and 0.8 and the two classes overlap widely
(a, AUC 0.91). The critic we freeze, at step 800, scoring the same rollouts, pushes incorrect
responses toward 0 and correct ones toward 1 (b, 0.94), and separates rollouts
from a converged policy just as clearly (c, 0.96). (d) Nine critic checkpoints at cumulative steps 100--900,
each averaged over ten actor distributions; the critic we freeze is the one at
step 800 (marked), and step 900 is a later checkpoint from the same run. Overall AUC also credits the critic
for recognizing which problems are hard; that is legitimate signal, but to
isolate the attempt we also measure within-problem AUC, which holds the problem
fixed and ranks attempts at it. It rises with critic training, holds at the
same level when restricted to rollouts that exhausted the generation budget, and
stays well above ranking by ``shorter is correct''. (e) Prefixes of $1{,}024$
rollouts (128 problems, 8 responses each) scored at increasing token lengths
against the final correctness of the full rollout. Overall AUC is high from the
first prefix; within-problem AUC starts near chance and rises as more of the
reasoning becomes visible, including on prefixes that have not yet produced an
answer.}
\label{fig:trust}
\end{figure}

\looseness=-2
\car asks one frozen critic to play three roles: a rollout-level reward, the
value baseline for generalized advantage estimation (GAE), and a forecaster of
success for unfinished prefixes. All three rest on the same quantity. Given a
prompt $x$, a policy generates a response $y=(y_1,\dots,y_T)$ and a verifier
returns a sparse terminal reward $\rstar(x,y)\in\{0,1\}$. PPO fits a critic
$V_\phi$ to empirical returns for GAE \citep{schulman2017ppo,schulman2016gae};
with $\gamma{=}\lambda{=}1$ the regression target at every prefix is the final
reward, so the ideal critic satisfies
\begin{equation}
V_\phi(x,y_{\le t}) \;\approx\;
\mathbb{E}_{\pi_\theta}\!\left[\rstar \mid x,y_{\le t}\right]
\;=\;\Pr\!\left(\text{success}\mid x,y_{\le t}\right)
\label{eq:identity}
\end{equation}
\citep{sutton2018reinforcement}. A trained critic only approximates this fixed
point, so what matters is how well it ranks rollouts, which we check offline.
We obtain such a critic from supervised PPO on mathematical reasoning: 500 steps
at an $8{,}192$-token budget and 300 more at a $5{,}120$-token cap, taking the
initial policy $\pi_{\theta_0}$ at cumulative step~700 and freezing the critic
$\vB$ at step~800 (Fig.~\ref{fig:provenance}; \S\ref{sec:setup}). Explained
variance, which costs nothing to monitor during training, rises from $-32$ at
initialization to about 0.55 by step~800 (Fig.~\ref{fig:provenance}a), and, as
the offline checks below show, it is a reliable indicator of critic quality.
A good critic separates rollout quality, and training is what makes it do so.
On rollouts from an early policy (cumulative PPO steps 100--200), the critic
checkpoint at cumulative step~200 separates correct from incorrect responses at
AUC 0.91; the critic we freeze, at step~800, reaches 0.94 on the same rollouts and
0.96 on rollouts from the converged policy (steps 700--800)
(Fig.~\ref{fig:trust}a--c). Scoring ten actor distributions ($16{,}384$ rollouts
each) with nine critic checkpoints from the same run (cumulative steps 100--900,
one past the frozen critic), overall AUC rises from
0.913 to 0.965 and within-problem AUC from 0.765 to 0.842, against 0.583 for
ranking by ``shorter is correct''; across the checkpoints, within-problem AUC
tracks explained variance ($r{=}0.91$; Fig.~\ref{fig:trust}d). The critic also
judges a rollout before it ends. Scoring prefixes of $1{,}024$ rollouts cut at
increasing lengths against the final correctness of each full rollout
(Fig.~\ref{fig:trust}e), overall AUC is already 0.958 at 256 tokens, largely
because the critic recognizes problem difficulty; ranking attempts at the same
problem starts near chance and rises with the visible reasoning, to 0.84 at $4{,}096$
tokens and 0.84--0.92 on prefixes that have not yet produced an answer. Rollouts
that exhaust the generation budget are ranked as well as the rest
(within-problem AUC 0.79--0.84), so this is not truncation detection.

\subsection{Freezing and calibrating the critic}
\label{sec:calib}

\looseness=-2
We freeze $\vB$ and give it two jobs at once. As the reward it supplies an
outcome score $v(x,y)=\vB(x,y_{\le T})$, where $T$ is the last \emph{observed}
response token---a rollout may end naturally or hit the generation cap, and
\car scores it either way. As a baseline it supplies the token-level values
$\vB(x,y_{\le t})$ that GAE needs. Both come out of the same forward pass the
old pipeline already ran, so the reward costs no additional model and no
additional pass.
Without a guard against length bias, the critic's raw terminal score cannot be
used directly as the reward. Length and correctness are mixed in the rollouts:
long responses are more often wrong, and part of the score's decline with length
reflects exactly that. But the score also falls with length when the outcome is
held fixed: among responses to the same problem with the same outcome, longer
ones score lower (correlation $-0.43$ on $16{,}384$ rollouts from the deployed
pair). That part is a bias, and the policy can exploit it by changing length
without changing correctness; a policy rewarded with the raw terminal score
simply stops reasoning (Appendix~\ref{app:reward}). Length
exploitation is a recurring problem in RL post-training, which is why many PPO-style recipes add explicit
length penalties or overlong-response shaping to the reward
\citep{singhal2023long,yu2025dapo,kimi2025k15}. The calibration below is our
guard. We fit a
debiasing term and a threshold on 512 on-policy rollouts from
$\pi_{\theta_0}$ and deploy
\begin{equation}
\rhat(x,y)\;=\;\mathbf{1}\!\left[\,v(x,y)-b\big(\ell(y)\big)>\tau\,\right],
\label{eq:car}
\end{equation}
where $\ell(y)$ is response length and $b(\ell)$ holds length-decile offsets
estimated within correctness class. Estimating within class is what separates the
bias from the signal: $b(\ell)$ captures how the score moves with length among
responses with the same outcome, and leaves the gap between correct and incorrect
responses in place. The threshold $\tau$ is the quantile of
those scores that matches their positive rate to the verifier-positive rate on
the calibration set. On our pair this gives $\tau=0.5215$ at 93.4\%
calibration accuracy. The cut must be fitted on the deployed pair: a critic can
rank well without its outputs being probabilities, and across the nine
checkpoints of \S\ref{sec:identity} the balanced-accuracy-optimal threshold
ranges from 0.462 to 1.082.
The two pieces do different work, and Appendix~\ref{app:reward} shows that both are
load-bearing. Under a continuous reward, whatever length dependence remains, in either direction, is something the policy
can climb; how strongly the score is debiased decides \emph{which way} it drifts.
The indicator decides \emph{whether} it drifts at all: once the reward is capped
at one, pushing a score higher above the cut earns nothing, and the exploit that
the continuous score invites has nowhere to go.
With $\gamma{=}\lambda{=}1$ and the reward of Eq.~(\ref{eq:car}), GAE collapses
to $A_t=\rhat-\vB(x,y_{\le t})$ before normalization, and the actor is updated
with the clipped PPO surrogate while $\phi$ is never touched
(Algorithm~\ref{alg:car}). An optional supervision fraction $p$ replaces $\rhat$
with $\rstar$ independently per trajectory, so $p{=}0$ is fully reward-free and
$p{=}1$ recovers supervised PPO with a trainable critic. Freezing also removes
the critic's optimizer states and backward pass from the step; we account for
what that saves in Appendix~\ref{app:cost}.

\begin{algorithm}[t]
\small
\caption{\car: calibrate once, then train with no labels}
\label{alg:car}
\begin{algorithmic}[1]
\Require actor $\pi_{\theta_0}$, critic $\vB$, prompts $\mathcal{D}$, group size $G$, supervision fraction $p$
\State Freeze $\vB$ \Comment{no critic optimizer, no critic backward pass}
\State Sample calibration rollouts $\mathcal{C}=\{(x,y,\rstar,\ell(y))\}$ from $\pi_{\theta_0}$
\State Fit length-decile offsets $b(\ell)$ within correctness class on $\mathcal{C}$
\State $q\gets\operatorname{mean}_{\mathcal{C}}(\rstar)$;\quad
       $\tau\gets Q_{1-q}\big(\vB(x,y_{\le T})-b(\ell(y))\big)$ on $\mathcal{C}$
\For{each RL step}
  \State Sample prompts from $\mathcal{D}$; generate $G$ responses each
  \State One forward pass of $\vB$ per response gives all token values; let $v_i=\vB(x_i,y_{i,\le T_i})$
  \State $\rhat_i\gets\mathbf{1}\!\left[v_i-b(\ell(y_i))>\tau\right]$
  \State $z_i\sim\operatorname{Bernoulli}(p)$;\quad $r_i\gets\rstar_i$ if $z_i=1$ else $\rhat_i$
  \State $A_{i,t}\gets r_i-\vB(x_i,y_{i,\le t})$; normalize advantages
  \State Update $\theta$ with the clipped PPO objective; leave $\phi$ unchanged
\EndFor
\end{algorithmic}
\end{algorithm}

\section{Experimental Setup}
\label{sec:setup}

\paragraph{Models and lineage.}\looseness=-1 
We fine-tune Qwen3-4B-Base \citep{yang2025qwen3} on 45k long chain-of-thought
solutions from OpenR1-Math-220k \citep{openr1} for three epochs and carry the
epoch-2 checkpoint forward. That checkpoint enters supervised PPO on
DAPO-Math-17k \citep{yu2025dapo}, run in the two consecutive phases described in
\S\ref{sec:provenance}: 500 steps at an $8{,}192$-token generation budget with
64 prompts per step, then 300 steps at a $5{,}120$-token training cap with 128.
We extract the initial policy $\pi_{\theta_0}$ at cumulative step~700 and the
critic $\vB$ at step~800. This is the only supervision anywhere in the pipeline,
and it amounts to roughly $0.56$M verifier-labeled trajectories.

\paragraph{What varies across runs, and what does not.}\looseness=-1 
Every run in the main comparison starts from that same $(\pi_{\theta_0},\vB)$
pair and draws the same prompts in the same nominal order under the same PPO
settings. Two things vary: where the reward comes from, and whether the critic
is updated. Supervised PPO takes verifier labels and trains its critic; \car
takes Eq.~(\ref{eq:car}) and freezes it. Table~\ref{tab:arms} lists the runs.
The three zero-label runs of the main comparison share their recorded seed
settings and repeat one condition.

\begin{table}[t]
\centering
\footnotesize
\setlength{\tabcolsep}{4pt}
\caption{\textbf{Training runs.} Shaded rows share the initial policy and the
initial critic and differ only in the reward; rows below the rule are \car
variants with parts of the calibration removed or changed, followed by the
$4{,}096$-token-cap pair added later (\S\ref{sec:results}). Steps are recorded run lengths,
which can exceed the 300-step comparison window. All runs train under a $5{,}120$-token cap unless the arm names another; the
raw-value run predates that protocol. Appendix~\ref{app:roster} maps each arm
to its log.}
\label{tab:arms}
\begin{tabular}{lccrl}
\toprule
\rowcolor{hdrbg}
Arm & Training reward & $p$ & Steps & Used for \\
\midrule
\rowcolor{carbg}
100\% labels (PPO) & verifier & 1.0 & 300 & label efficiency \\
\rowcolor{carbg}
\car, 50\% labels & verifier or Eq.~(\ref{eq:car}) & 0.5 & 330 & label efficiency \\
\rowcolor{carbg}
\car, 0\% labels ($3\times$) & Eq.~(\ref{eq:car}) & 0 & 310/324/378 & main result \\
\midrule
Raw value & $v$, no calibration & 0 & 256 & App.~\ref{app:reward} \\
Threshold only & $\mathbf{1}[v>0.6]$ & 0 & 100 & App.~\ref{app:reward} \\
Continuous $v$ ($4\times$) & $v-b(\ell)$, no threshold & 0 & 87/141/220/257 & App.~\ref{app:reward} \\
Continuous $v$, GRPO & same, group-relative & 0 & 161 & App.~\ref{app:reward} \\
$1{,}024$-token cap & $v$ & 0 & 144 & rollout budget \\
\midrule
\car, $4{,}096$-token cap & Eq.~(\ref{eq:car}), refitted & 0 & 310 & rollout budget \\
PPO, $4{,}096$-token cap & verifier & 1.0 & 310 & rollout budget \\
\bottomrule
\end{tabular}
\end{table}

\paragraph{Optimization settings, shared by \car and the supervised baseline.}\looseness=-1 
\car updates the policy with the clipped surrogate objective of PPO
\citep{schulman2017ppo}; what it replaces is the reward and the trained critic,
not the optimizer. Both \car and supervised PPO therefore run under identical
settings, in verl \citep{sheng2024hybridflow}, with 128 prompts and 8 responses
per prompt, so every step consumes $1{,}024$ rollouts. Prompts are capped at
$2{,}048$ tokens, training generations at $5{,}120$ and validation generations
at $12{,}288$. Sampling is at temperature 1.0, the clip ratio is 0.2,
$\gamma{=}\lambda{=}1$, the actor learning rate is $10^{-6}$, and the critic
learning rate is $10^{-5}$ in the supervised baseline, the only arm that trains
one. No KL penalty is applied,
either in the reward or as a loss. Under this protocol the three zero-label runs
truncate 26.5--28.5\% of training generations at the $5{,}120$ cap, while
validation at $12{,}288$ tokens is capped for only 1.1--1.8\%. Critic-based RL is widely held to be unstable on long chains of thought \citep{yuan2025vcppo,yue2025vapo}, and our own
earlier runs reproduced the symptom: with 2 to 16 inner updates over batches of
256 to $1{,}024$ rollouts, supervised PPO and reward-free runs alike failed in
two opposite directions, policy entropy either collapsing toward zero or
drifting upward without settling. The cause was the update, not the critic.
Keeping each policy update small and low in variance---one gradient step per
batch, with the mini-batch equal to the full batch and no inner loop over the
collected rollouts---makes both failures recede, and under that single rule
supervised PPO and \car alike train stably for hundreds of steps
(Appendix~\ref{app:inner}). This is the evidence
behind contribution~(i), and the rule is used in every run reported below.

\paragraph{Calibration.}\looseness=-1 
The debiasing term and threshold of Eq.~(\ref{eq:car}) are fitted once, on
512 on-policy rollouts drawn from $\pi_{\theta_0}$ at the training cap, and then
never refitted: $b(\ell)$ from length-decile offsets within correctness class,
and $\tau=0.5215$ from matching the positive rate of the corrected scores to the
verifier-positive rate on that sample, at which the binary rule agrees with the
verifier on 93.4\% of it.

\paragraph{Evaluation.}\looseness=-1 
Every ten steps we evaluate AIME 2026 (30 problems) and AMC 2023 (40 problems)
with 8 samples per problem at temperature 0.7 and a $12{,}288$-token budget, and
report their macro average, which weights the two suites equally. Training-time
pass@$k$ uses the plug-in estimator
$\mathbb{E}_{\text{problem}}[1-(1-\hat p)^k]$ on the sampled solve rate $\hat
p$. The principal comparison window is the first 300 RL steps; runs recorded
past that point also contribute to peak and late-training summaries, with their
windows stated. Step-300 checkpoints additionally receive a separate $n{=}16$
evaluation on AIME 2025, AIME 2026, AMC 2023 and GPQA-Diamond
\citep{rein2024gpqa}; GPQA is the out-of-domain probe.

\section{Results}
\label{sec:results}

\looseness=-1
Four questions decide whether a frozen critic is usable as the reward: whether
training stays stable, whether the policy gets as far as it
would with a verifier, whether
a fixed reward survives the policy that is optimizing against it, and whether
the reward can be assigned before a rollout finishes.
\textbf{(1)~Binarization trades a higher ceiling for stability.} 
Used as a continuous reward, the same critic climbs above fully supervised
PPO---29.2 against 25.0 on AIME 2026 pass@1, the highest AIME score in this
paper---showing how much signal a pretrained critic carries
(Fig.~\ref{fig:aime}a--b, Appendix~\ref{app:eval}). The gain does not hold: every continuous arm declines
later in training, and response length drifts with it, shorter or longer
depending on how strongly the score is debiased, as the policy exploits whatever
length bias the continuous reward still carries. Binarizing forfeits that ceiling and keeps training stable, which
is why \car deploys it (Appendix~\ref{app:reward}).
\textbf{(2)~Training stays stable.} 
Under the single-update rule of \S\ref{sec:setup}, none of the three zero-label
runs degenerates over its 310--378 steps. Smoothed response length stays within
13\% of its starting value, where the uncalibrated raw score of
Appendix~\ref{app:reward} shortens responses by 73\% within 100 steps; policy entropy
declines gradually from 0.32 to 0.21--0.22 by step~300, against 0.25 for
supervised PPO, neither collapsing nor drifting upward; and held-out accuracy
shows no late collapse (Table~\ref{tab:dynamics}). A frozen reward with no
verifier in the loop therefore trains as steadily as the supervised baseline.

\begin{figure}[tb]
\centering
\includegraphics[width=\textwidth]{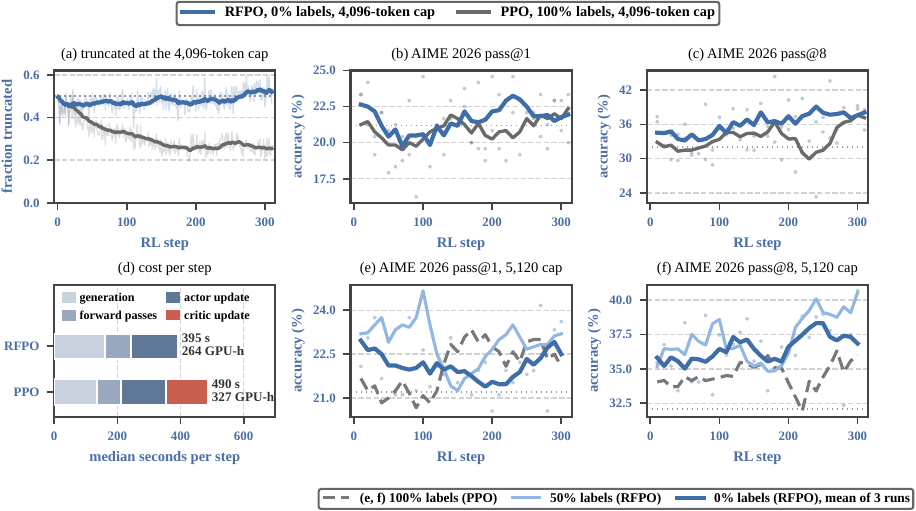}
\caption{\textbf{Training with half of every batch unfinished, and the main
result.} (a--d) \car (zero labels) and supervised PPO, both under a
$4{,}096$-token cap: (a) share of training rollouts truncated; (b,\,c) AIME 2026
validation; (d) median seconds per step by component on the same node, with
GPU-hours for 300 steps. (e,\,f) The main runs at the $5{,}120$-token cap: AIME
2026 validation of supervised PPO, the half-labeled arm and the mean of the three
zero-label runs (Table~\ref{tab:dynamics}). Validation is at $12{,}288$ tokens
(dots: one evaluation every 10 steps, for the zero-label mean the mean of the
three runs' evaluations at that step; lines: centered moving averages over five
consecutive evaluations, $\pm 20$ steps; dotted: initial policy); AMC 2023 in Table~\ref{tab:cap4096}, macro-averages in
Figure~\ref{fig:overview}.}
\label{fig:cap4096}
\end{figure}

\looseness=-1
\textbf{(3)~\car reaches supervised PPO.} 
Table~\ref{tab:main} reports the separate 16-sample evaluation at RL step 300
of the $5{,}120$-token-cap runs.
Averaged over four suites, the zero-label checkpoint reaches 41.1\% pass@1
against 41.8\% for supervised PPO, 41.3\% for the half-labeled arm and 40.0\%
for the shared initial policy: relative to supervised PPO it is 0.2 points
higher on AIME 2025, tied on AIME 2026, 0.7 higher on GPQA-Diamond and 3.6
lower on AMC 2023. Training-time validation tells the same story across all
three zero-label runs (Fig.~\ref{fig:cap4096}e--f; per-run numbers in
Table~\ref{tab:dynamics}): they peak at 25.7\% on AIME 2026 pass@1 against 25.0\%
for supervised PPO, and at 76.6\% on AMC 2023 pass@1 against 78.1\%. The
differences are small: level or ahead on AIME and GPQA, and 1.5--3.6 points
behind on AMC 2023 pass@1, a 40-problem suite on which one problem is worth 2.5
points. With no verifier anywhere in its optimization loop, \car performs on
par with fully supervised PPO.
\textbf{(4)~A frozen reward survives drift, and partial labels can guard it.} 
The policy moves during training while the reward stays fixed. Re-applying the frozen critic to each arm's own rollouts across
training, its ranking of terminal correctness shows no significant trend over
310 steps on any healthy arm (AUC slopes between $-0.006$ and $+0.002$ per 100
steps, all $|t|{<}0.8$). The only significant decline anywhere in our data is
the continuous arm of Fig.~\ref{fig:aime}a--b ($-0.010$ per 100 steps,
$t{=}{-}2.89$), whose rollouts almost stop finishing: the critic did not get
worse, it was handed a population it was never fitted on, and binarization is
what keeps the policy out of that region (Appendix~\ref{app:drift}). Where more
protection is wanted, the supervision fraction offers it: mixing the verifier
into a random half of trajectories performs on par with both extremes (last-ten
AIME pass@8 39.1\% against 34.4\% for supervised PPO and 37.2\% for the
zero-label runs), so the fraction is a continuous dial rather than a switch.

\begin{table}[t]
\centering
\setlength{\tabcolsep}{5pt}
\footnotesize
\caption{\textbf{Separate evaluation after 300 RL steps.} Pass@1 (\%) at 16
samples, temperature 0.7 and a $12{,}288$-token budget; Avg is the mean over the
four suites. Each RL row is one
checkpoint; the zero-label row covers one of the three runs, the other two being
assessed through training-time validation. $^{\dag}$Base is evaluated as plain
completion, without the think template.}
\label{tab:main}
\begin{tabular}{lccccc}
\toprule
\rowcolor{hdrbg}
Model & AIME25 & AIME26 & AMC23 & GPQA & Avg \\
\midrule
Qwen3-4B-Base$^{\dag}$ & 3.1 & 3.5 & 25.9 & 33.0 & 16.4 \\
\ + SFT (45k CoT) & 22.3 & 20.8 & 63.4 & 35.7 & 35.5 \\
Initial policy $\pi_{\theta_0}$ & 22.7 & 21.0 & 71.4 & 44.9 & 40.0 \\
\midrule
\ \ + PPO, 100\% labels & \underline{25.2} & \textbf{21.7} & \textbf{73.6} & 46.6 & \textbf{41.8} \\
\rowcolor{carbg}
\ \ + \car, 50\% labels & 22.7 & \underline{21.2} & \underline{72.8} & \textbf{48.5} & \underline{41.3} \\
\rowcolor{carbg}
\ \ + \car, 0\% labels & \textbf{25.4} & \textbf{21.7} & 70.0 & \underline{47.3} & 41.1 \\
\bottomrule
\end{tabular}
\end{table}

\begin{table}[t]
\centering
\footnotesize
\setlength{\tabcolsep}{4pt}
\caption{\looseness=-1 \textbf{Peak and final validation across every run.} Each cell is
peak\,{\scriptsize(step)}\,/\,final over each run's recorded length. The shaded
rows summarize the three zero-label runs. All rows train under the
$5{,}120$-token cap except the last two, the $4{,}096$-token-cap pair added later (Table~\ref{tab:cap4096}).}
\label{tab:dynamics}
\begin{tabular}{lrcccc}
\toprule
\rowcolor{hdrbg}
& & \multicolumn{2}{c}{AIME 2026} & \multicolumn{2}{c}{AMC 2023} \\
\cmidrule(lr){3-4}\cmidrule(lr){5-6}
\rowcolor{hdrbg}
Arm & Steps & pass@1 & pass@8 & pass@1 & pass@8 \\
\midrule
Initial policy $\pi_{\theta_0}$ & 0 & 21.2 & 32.1 & 73.6 & 89.5 \\
\midrule
PPO, 100\% labels & 300 & 25.0\,{\scriptsize(190)}\,/\,21.7 & 41.7\,{\scriptsize(250)}\,/\,34.0 & \textbf{78.1}\,{\scriptsize(120)}\,/\,71.9 & 93.2\,{\scriptsize(210)}\,/\,89.0 \\
\car, 50\% labels & 330 & 25.8\,{\scriptsize(80)}\,/\,22.5 & 42.9\,{\scriptsize(300)}\,/\,39.3 & \textbf{78.1}\,{\scriptsize(290)}\,/\,71.2 & 94.1\,{\scriptsize(290)}\,/\,89.0 \\
\midrule
\car, 0\% labels, run 1 & 310 & 24.6\,{\scriptsize(220)}\,/\,20.0 & 42.7\,{\scriptsize(220)}\,/\,33.0 & 76.2\,{\scriptsize(290)}\,/\,73.1 & 93.4\,{\scriptsize(290)}\,/\,85.7 \\
\car, 0\% labels, run 2 & 324 & \textbf{26.2}\,{\scriptsize(30)}\,/\,23.8 & 43.7\,{\scriptsize(130)}\,/\,39.3 & 75.9\,{\scriptsize(40)}\,/\,72.2 & 93.8\,{\scriptsize(180)}\,/\,90.9 \\
\car, 0\% labels, run 3 & 378 & \textbf{26.2}\,{\scriptsize(270)}\,/\,20.0 & \textbf{44.5}\,{\scriptsize(140)}\,/\,31.5 & 77.5\,{\scriptsize(30)}\,/\,71.6 & \textbf{93.6}\,{\scriptsize(30)}\,/\,91.0 \\
\rowcolor{gray!12}
\textbf{Reward-free} {\scriptsize(mean)} & & \textbf{25.7}\,/\,21.2 & \textbf{43.7}\,/\,34.6 & 76.6\,/\,72.3 & 93.6\,/\,89.2 \\
\rowcolor{gray!12}
\quad {\scriptsize\emph{s.d.\ over the three runs}} & & {\scriptsize 1.0\,/\,2.2} & {\scriptsize 0.9\,/\,4.2} & {\scriptsize 0.8\,/\,0.8} & {\scriptsize 0.2\,/\,3.0} \\
\midrule
PPO, 100\% labels, $4{,}096$ cap & 310 & 24.6\,{\scriptsize(200)}\,/\,23.3 & 40.3\,{\scriptsize(200)}\,/\,35.0 & 75.6\,{\scriptsize(20)}\,/\,71.6 & 93.0\,{\scriptsize(270)}\,/\,86.3 \\
\car, 0\% labels, $4{,}096$ cap & 310 & 24.6\,{\scriptsize(100)}\,/\,20.0 & 44.4\,{\scriptsize(180)}\,/\,38.6 & 76.9\,{\scriptsize(60)}\,/\,72.5 & 93.6\,{\scriptsize(60)}\,/\,83.7 \\
\bottomrule
\end{tabular}
\end{table}

\begin{table}[tb]
\centering
\footnotesize
\setlength{\tabcolsep}{4pt}
\caption{\textbf{4,096-token cap, per benchmark.} Validation at $12{,}288$
tokens (\%), each cell the mean over the evaluations at steps 10--300 / the peak
in that window; GPU-hours for 300 steps at the median step time. The first two
rows ran on the same node; the last row is supervised PPO at the full
$5{,}120$-token cap (Table~\ref{tab:dynamics}), for reference.}
\label{tab:cap4096}
\begin{tabular}{lccccc}
\toprule
\rowcolor{hdrbg}
& \multicolumn{2}{c}{AIME 2026} & \multicolumn{2}{c}{AMC 2023} & \\
\cmidrule(lr){2-3}\cmidrule(lr){4-5}
\rowcolor{hdrbg}
Arm & pass@1 & pass@8 & pass@1 & pass@8 & GPU-h \\
\midrule
\rowcolor{carbg}
\car, 0\% labels, $4{,}096$ cap & \textbf{21.6} / \textbf{24.6} & \textbf{36.1} / \textbf{44.4} & \textbf{73.5} / \textbf{76.9} & \textbf{90.1} / \textbf{93.6} & \textbf{264} \\
PPO, 100\% labels, $4{,}096$ cap & 20.9 / \textbf{24.6} & 33.5 / 40.3 & 72.4 / 75.6 & 89.5 / 93.0 & 327 \\
\midrule
PPO, 100\% labels, $5{,}120$ cap & 22.1 / 25.0 & 34.5 / 41.7 & 74.4 / 78.1 & 89.8 / 93.2 & 388 \\
\bottomrule
\end{tabular}
\end{table}

\looseness=-1
\textbf{(5)~The reward does not wait for the rollout to end.} 
Under the deployed protocol 26.5--28.5\% of training generations hit the
$5{,}120$-token cap, and \car scores them at the last observed token exactly like
any other---so more than a quarter of every batch is already being rewarded
without a finished answer. Varying the generation cap and really regenerating
under it shows how much that costs. At caps of $5{,}120$, $8{,}192$ and
$12{,}288$ tokens the fraction of rollouts that finish is 0.735, 0.941 and
0.967, an eightfold change in the unfinished population, while the reward's
precision stays at 0.894--0.896 and its explained variance at 0.66--0.70.
Scoring unfinished rollouts costs essentially nothing over this range, and
training confirms it (Figs.~\ref{fig:overview} and~\ref{fig:cap4096}). Under a $4{,}096$-token cap, with
$b(\ell)$ refitted so that short and long responses of the same outcome again
score alike, about half of every batch stays unfinished for 310 steps, yet \car's
mean validation (47.6\%) is above that of supervised PPO trained under the same
cap (46.6\%), on all four benchmark metrics (Table~\ref{tab:cap4096}), and level with our $5{,}120$-token runs (47.2--47.4\%), for 264
rather than 327 GPU-hours over 300 steps; supervised PPO with the full
$5{,}120$-token cap reaches 48.3\% for 388 (App.~\ref{app:budget}).
The approach has a floor: rollouts that stop resembling anything the critic was
fitted on. Within a problem, the critic ranks $1{,}024$-token prefixes at an AUC
of only 0.60, against 0.84 from $4{,}096$ tokens on (\S\ref{sec:trust},
Table~\ref{tab:prefixfull}); under a real $1{,}024$-token cap only 0.7\% of
rollouts finish and precision collapses to 0.213. A run trained there raises its
training solve rate from 5.5\% to 19.3\% while validation falls from 49.1\% to
39.1\% (Appendix~\ref{app:budget}).
\textbf{(6)~Cost falls sharply.} Freezing the critic deletes one of the two trained modules
from every step, and the reward needs no forward pass of its own: it is read off
the value pass PPO already performs. On matched batches a step takes 421 rather
than 582\,s ($-28\%$) and peak reserved memory falls by 9.4\,GB per GPU
(Appendix~\ref{app:cost}, Figure~\ref{fig:cost}), and
Figure~\ref{fig:cap4096}d shows the same cut at $4{,}096$ tokens. The savings apply to the RL
stage; obtaining and calibrating the critic is a separate, one-off preparation
cost.

\section{Limitations}
\label{sec:limitations}

All our experiments use a single model family at 4B parameters on mathematical
reasoning. We have not yet tried larger models or non-reasoning tasks, because
the experiments are expensive: a single 300-step RL run takes 264--388 GPU-hours
on eight A100-80GB GPUs (Appendix~\ref{app:cost}). Extending \car to larger
models and to other tasks is left to future work.

\section{Conclusion}
\label{sec:conclusion}

\looseness=-1
Reasoning RL has been removing the value network to buy stability and memory, on
the tacit assumption that it is worth only what it contributes as a baseline. We
have argued the opposite. The instability behind its removal comes from the
update recipe rather than from the network, and a well-pretrained critic carries
far more than a baseline needs: it estimates how likely a rollout is to succeed,
and it does so before the rollout ends.
\car puts that signal to work. A single frozen critic serves at once as the
rollout-level reward, the GAE baseline and a forecaster for unfinished prefixes.
Used as a continuous reward it lifts accuracy above fully supervised PPO, which
shows how much the critic knows; binarized, it gives up that ceiling in exchange
for stable training and matches supervised PPO with no verifier in the loop and a
lower per-step cost. Because its forecasts score truncated rollouts with accuracy
comparable to complete ones, training need not wait for every trajectory to
finish.
That last property is where we think this leads. As reasoning traces lengthen and
agentic episodes stretch, waiting for the outcome buys less and less for what it
costs, and the value network is the one component of the standard recipe that
speaks before the outcome arrives. The question we would put back on the table is not
whether a critic is affordable, but how much of what it already knows the
current recipe throws away. \car is our answer: it keeps that knowledge and turns
it into a reward-free, compute-efficient method for LLM post-training.

\section*{Reproducibility Statement}

Every number here is produced from artifacts included with the submission.
\S\ref{sec:setup} and Appendix~\ref{app:roster} give the model lineage, the PPO
configuration and the mapping from each arm in Table~\ref{tab:arms} to its log;
\S\ref{sec:calib} and Table~\ref{tab:lenbaseline} give the calibration rule of
Eq.~(\ref{eq:car}) with its fitted offsets and threshold; Appendices
\ref{app:eval}--\ref{app:critic} give the evaluation estimators and the
critic-side measurements. The archive contains the training log of every run
reported, the frozen critic's calibration artifacts, the per-cell scoring
summaries behind the measurement plane and the prefix scan, the benchmark
outputs, and one script that regenerates every figure and table from them.
Regenerating the figures needs only the archive; recomputing the critic-side
measurements additionally needs the rollout archives and the frozen checkpoint,
whose sizes and original locations the file index lists.

\section*{AI Use Statement}

We used generative AI tools to polish the writing, draft parts of the text, and
assist with computing summary statistics from training logs and plotting figures.
All experiments were designed, implemented and run by the authors, who made all
analysis decisions and verified every reported number against the recorded logs.
The authors take full responsibility for the content of this paper.

\bibliographystyle{iclr2027_conference}
\bibliography{refs}

\clearpage
\appendix
\section{Training recipes}
\label{app:recipes}

This appendix gives everything needed to rebuild the pipeline of
\S\ref{sec:setup}: the SFT model, the supervised run that yields the critic and
the initial policy, the RL stage shared by \car and supervised PPO, and the
calibration fit. Every run uses verl \citep{sheng2024hybridflow}.

\begin{figure}[h]
\centering
\includegraphics[width=\textwidth]{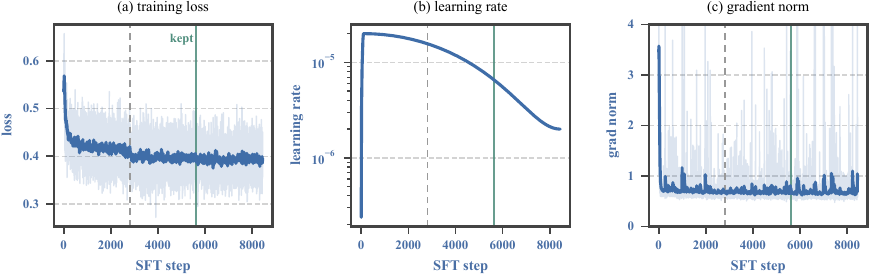}
\caption{\textbf{Supervised fine-tuning.} Loss, learning rate and gradient norm
over three epochs; the dashed line marks the first epoch boundary and the green
line the epoch-2 checkpoint carried into RL.}
\label{fig:sft}
\end{figure}

\paragraph{Supervised fine-tuning.}
We fine-tune Qwen3-4B-Base on a 45k subset of OpenR1-Math-220k \citep{openr1},
keeping solutions that close their reasoning block exactly once and end in a
boxed answer (99\% of the sampled pool pass). Training runs for three epochs of
$2{,}812$ steps at a global batch of 16 sequences and a maximum length of
$32{,}768$ tokens, with AdamW, learning rate $2\times10^{-5}$, cosine decay to
10\% after a 1\% warmup, weight decay 0.1, bf16, FSDP and gradient
checkpointing on $4\times4$ A100 GPUs. We keep the epoch-2 checkpoint: on AIME
2026 ($n{=}4$, $16{,}384$ tokens) the three epochs score 20.0\%, 25.0\% and
20.8\%, and the loss is flat after the first epoch
(Figure~\ref{fig:sft}). One detail matters for reproduction: under
single-message tokenization, the default Qwen3 chat template silently strips the
\texttt{<think>} block, removing most of the loss target; with the unpatched
template the same run reaches only 5\% on AIME, so we patch the template to keep
the reasoning.

\paragraph{Critic pretraining.}
The critic and the initial policy come out of one supervised PPO run on
DAPO-Math-17k \citep{yu2025dapo} in two phases (Table~\ref{tab:hypers}). Phase~1
starts both actor and critic from the SFT weights, the critic with a freshly
initialized scalar value head, and trains the critic alone for the first 30
steps. Phase~2 resumes both from phase-1 step 500 and changes three things: the
batch doubles to 128 prompts, the training cap drops to $5{,}120$ tokens, and
validation moves to $12{,}288$. We extract the actor at phase-2 step 200
(cumulative 700) and the critic at step 300 (cumulative 800), so the critic has
fitted 100 further steps of an improving policy. Over its 800 steps the critic
sees $563{,}200$ verifier-labeled rollouts. Table~\ref{tab:stages} shows what
the two supervised stages do to the model.

\begin{table}[h]
\centering
\footnotesize
\setlength{\tabcolsep}{3.8pt}
\caption{\textbf{What the supervised stages install.} $n{=}16$ samples at
temperature 0.7 with a $12{,}288$-token budget; mean length and cap hit are
averaged over the four suites. SFT buys accuracy with runaway generation, and
the PPO run that trains the critic brings length back down while adding accuracy,
most on GPQA-Diamond. $^\dagger$Evaluated without the chat template and without
thinking mode.}
\label{tab:stages}
\begin{tabular}{lcccccc}
\toprule
\rowcolor{hdrbg}
& \multicolumn{4}{c}{pass@1 (\%)} & & \\
\rowcolor{hdrbg}
Model & AIME 2025 & AIME 2026 & AMC 2023 & GPQA-D & Mean length & Cap hit (\%) \\
\midrule
Qwen3-4B-Base$^\dagger$ & 3.1 & 3.5 & 25.9 & 33.0 & 950 & 3.5 \\
+ SFT & 22.3 & 20.8 & 63.4 & 35.7 & 8{,}719 & 48.5 \\
\rowcolor{carbg}
+ critic pretraining ($\pi_{\theta_0}$) & 22.7 & 21.0 & 71.4 & 44.9 & 3{,}793 & 2.3 \\
\bottomrule
\end{tabular}
\end{table}

\paragraph{RL stage.}
Table~\ref{tab:hypers} lists the settings. \car and supervised PPO share every
one of them; they differ only in the reward and in whether the critic is
updated. The DAPO overlong penalty present in the reward manager is configured
to start at $8{,}192$ tokens and so never fires under the $5{,}120$-token training
cap; validation accuracy is computed from the verifier score alone.

\begin{table}[h]
\centering
\footnotesize
\setlength{\tabcolsep}{3.5pt}
\caption{\textbf{Hyperparameters.} Rows above the rule change between stages;
rows below are shared by every stage.}
\label{tab:hypers}
\begin{tabular}{llll}
\toprule
\rowcolor{hdrbg}
Setting & Critic pretraining, ph.\ 1 & Phase 2 & RL stage (PPO and \car) \\
\midrule
Actor init & SFT & phase-1 step 500 & $\pi_{\theta_0}$ (cumulative 700) \\
Critic init & SFT, new value head & phase-1 step 500 & $\vB$ (cumulative 800) \\
Critic & trained & trained & PPO: trained; \car: frozen \\
Reward & verifier & verifier & PPO: verifier; \car: Eq.~(\ref{eq:car}) \\
Steps & 500 & 300 & 300 compared (runs to 378) \\
Prompts $\times$ responses & $64\times8$ & $128\times8$ & $128\times8$ \\
Train / val.\ generation cap & $8{,}192$ / $8{,}192$ & $5{,}120$ / $12{,}288$ & $5{,}120$ / $12{,}288$ \\
\midrule
Updates per batch & \multicolumn{3}{l}{one (mini-batch = batch, one epoch; App.~\ref{app:inner})} \\
Prompt cap & \multicolumn{3}{l}{$2{,}048$ tokens} \\
Rollout sampling & \multicolumn{3}{l}{temperature 1.0, top-$p$ 1.0, vLLM with tensor parallelism 2} \\
Validation & \multicolumn{3}{l}{every 10 steps on AIME 2026 and AMC 2023, 8 samples at temp.\ 0.7} \\
Advantage & \multicolumn{3}{l}{GAE, $\gamma=\lambda=1$, whitened over the batch} \\
Policy loss & \multicolumn{3}{l}{clipped surrogate, $\epsilon=0.2$ both sides, token-mean; no entropy or KL term} \\
Value loss & \multicolumn{3}{l}{clipped at 0.5 (trained critics only)} \\
Learning rate & \multicolumn{3}{l}{actor $10^{-6}$, critic $10^{-5}$; constant after 10 warmup steps} \\
Optimizer & \multicolumn{3}{l}{AdamW, $\beta=(0.9,0.999)$, weight decay 0.1, gradient clipping 1.0} \\
Systems & \multicolumn{3}{l}{bf16, FSDP, sequence parallelism 2, gradient checkpointing} \\
Hardware & \multicolumn{3}{l}{one node, $8\times$A100-80GB} \\
\bottomrule
\end{tabular}
\end{table}

\paragraph{Calibration fit.}
The calibration sample is the 512 rollouts that $\pi_{\theta_0}$ generated at its
own training step (phase-2 step 200, $5{,}120$-token cap), scored by $\vB$ with
the prompt in context; 53.5\% are verifier-correct. Rollouts are split into
length deciles. Within each decile we compute, separately for correct and for
incorrect rollouts, the mean score minus that class's overall mean, and average
the two deviations; this is $b(\ell)$ (Table~\ref{tab:lenbaseline}). Working
within correctness class removes the length trend inside each class while leaving
the genuine fact that long responses are more often wrong in the class mix. The
threshold $\tau$ is the $(1-0.535)$ quantile of the corrected scores
$v-b(\ell)$, which gives $\tau=0.5215$ and 93.4\% agreement with the verifier on
this sample. Both are fitted once and never refitted.

\begin{table}[h]
\centering
\footnotesize
\setlength{\tabcolsep}{2.4pt}
\caption{\textbf{Deployed length baseline.} Lengths outside the fitted range take
the nearest decile's offset. The last decile holds the rollouts that reach
the cap.}
\label{tab:lenbaseline}
\begin{tabular}{lcccccccccc}
\toprule
\rowcolor{hdrbg}
Decile & 1 & 2 & 3 & 4 & 5 & 6 & 7 & 8 & 9 & 10 \\
\midrule
Upper edge & 1{,}735 & 2{,}137 & 2{,}455 & 2{,}782 & 3{,}150 & 3{,}556 & 3{,}884 & 4{,}469 & 5{,}079 & 5{,}120 \\
$b(\ell)$ & $+0.071$ & $+0.087$ & $+0.100$ & $+0.065$ & $+0.083$ & $+0.063$ & $+0.024$ & $-0.070$ & $-0.104$ & $-0.224$ \\
\bottomrule
\end{tabular}
\end{table}

\section{Inner updates per batch and training stability}
\label{app:inner}

Before settling on the recipe of \S\ref{sec:setup}, we trained supervised PPO and
reward-free variants the way PPO is usually run, with several optimizer steps on
each batch of rollouts. This appendix collects that record.

\paragraph{What was run.}
From all our logged PPO and \car runs on math reasoning we keep the 74 that
reached at least 60 steps; the rest, including all three runs with 16 inner
updates, stopped earlier. Inner updates count the optimizer steps taken on one
batch (batch size over mini-batch size, times PPO epochs), and runs are grouped
by inner updates and rollouts per step. A run \emph{collapses} if its policy
entropy falls below 0.35 times its starting value (the mean of its first ten
steps) and \emph{diverges} if entropy rises above 1.8 times that value; every
other run is \emph{stable}. The criterion concerns optimization only and does not
ask whether accuracy holds: the continuous-score arms of
Appendix~\ref{app:reward}, whose accuracy peaks and then declines, count as
stable here.

\begin{figure}[t]
\centering
\includegraphics[width=\textwidth]{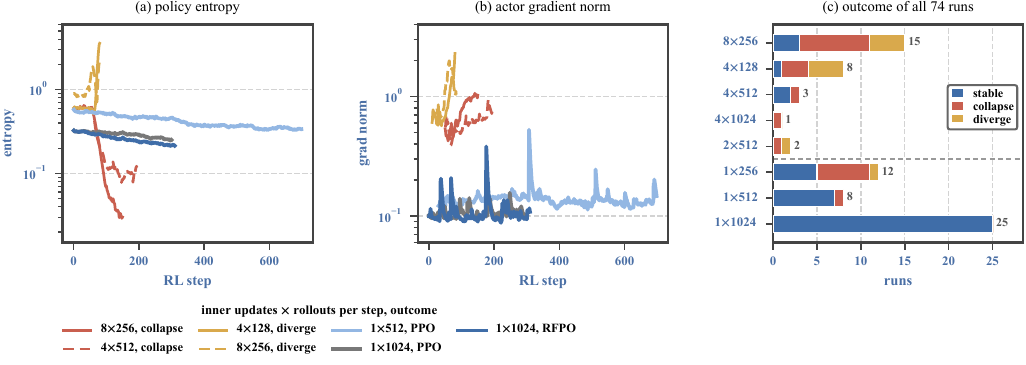}
\caption{\textbf{Inner updates and training stability.} (a,b) Policy entropy and
actor gradient norm for representative runs, labeled by inner updates $\times$
rollouts per step; multi-update runs collapse (red) or diverge (yellow) within 200
steps, while single-update runs stay flat. (c) Outcome of all 74 runs by
configuration; the dashed line separates multi-update from single-update groups.
Runs differ in initial policy, response cap, upper clip ratio and reward; see
text.}
\label{fig:stability}
\end{figure}

\paragraph{What we found.}
With more than one inner update, 23 of 29 runs failed, in both directions and at
every batch size we tried; at 8 updates over 256 rollouts, 12 of 15 did. Their
median gradient norms over the last ten steps range from 0.27 to 0.89 by group.
With a single update per batch, 8 of 45 runs failed, all seven supervised PPO runs
were stable, and all 25 runs at $1{,}024$ rollouts, supervised and reward-free
alike, stayed stable with a median final gradient norm of 0.10
(Figure~\ref{fig:stability}). A single update also takes the clip out of play:
the gradient is taken at the policy that generated the rollouts, the importance
ratio is one, and the clip fraction and PPO KL are exactly zero in every such
run. Stability under this rule therefore does not come from the trust region; it
comes from keeping each step small and averaged over many rollouts.

\paragraph{The failures that remain.}
All eight single-update failures ran at 256 or 512 rollouts, and none used the
calibrated reward of Eq.~(\ref{eq:car}): three withheld outcome labels with no
calibrated replacement, four rewarded the raw critic score, and one thresholded
it at a fixed 0.7 with no length baseline. They combine a smaller batch with a
reward the final recipe dropped, and do not implicate the single-update rule.

\paragraph{Other differences between runs.}
The runs differ in more than the update: in initial policy, response cap
($8{,}192$ to $20{,}480$ tokens), upper clip ratio (0.28 in most early
multi-update runs) and reward. Two checks bound this. The multi-update failures
span three different starting checkpoints, and all eight
multi-update runs that used the symmetric clip of 0.2 of the final recipe failed
as well, so neither initialization nor the upper clip explains them. The reverse
confound remains, since the $1{\times}1{,}024$ group is also where the final reward
lives. The record supports what the paper uses: no multi-update configuration we tried trained
reliably, and the single-update rule at $1{,}024$ rollouts did not fail in 25 runs.

\section{Evaluation protocol and extended results}
\label{app:eval}
\label{app:benchmarks}

\begin{figure}[t]
\centering
\includegraphics[width=\textwidth]{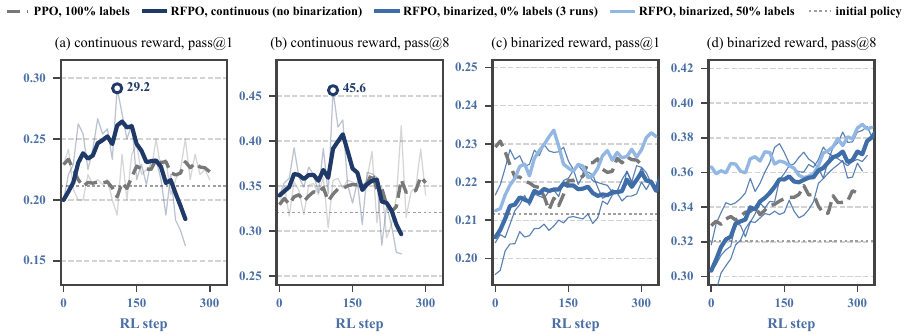}
\caption{\textbf{Held-out AIME 2026 accuracy under the two rewards.} (a--b) \car
with the critic's debiased continuous score as the reward (no binarization)
against PPO trained on every label. Faint traces are raw validation and solid
lines smoothed trends; the marker is the highest single evaluation. The
continuous reward reaches 29.2 pass@1 and 45.6 pass@8, above anything supervised
PPO attains, and then declines below it. (c--d) Deployed, binarized \car: the
three zero-label runs are drawn individually (thin) with their mean (thick),
since each labeled arm is a single run, and all curves are smoothed. Binarized
\car tracks supervised PPO throughout. Dotted lines mark the shared initial
policy; smoothed curves understate single-evaluation peaks, which are listed per
run in Table~\ref{tab:dynamics}. The other continuous settings are in
Appendix~\ref{app:reward}, and AMC 2023 results in Appendix~\ref{app:benchmarks}.}
\label{fig:aime}\label{fig:ceiling}\label{fig:main}
\end{figure}

\paragraph{Protocol.}
Every ten steps we sample 8 responses per problem on AIME 2026 (30 problems) and
AMC 2023 (40 problems) at temperature 0.7, top-$p$ 1.0 and a $12{,}288$-token
budget. The budget is deliberately larger than the $5{,}120$-token training cap,
so validation measures what the policy can do with room to reason rather than
the training constraint. Correctness comes from the same rule-based verifier used
for training: the final boxed expression is matched against the reference answer
after normalization. No model-based judge is used anywhere in this paper. The
step-300 checkpoints are evaluated separately with 16 samples per problem on four
suites; GPQA-Diamond (198 multiple-choice questions) is the out-of-domain suite.

\paragraph{Estimators.}
With $\hat p$ the fraction of a problem's $n$ samples that are correct, pass@1 is
$\mathbb{E}_{\text{problem}}[\hat p]$ and training-time pass@$k$ is the plug-in
$\mathbb{E}_{\text{problem}}[1-(1-\hat p)^k]$. The plug-in is biased at $k{=}n$:
a problem solved once in eight contributes $1-(7/8)^8\approx0.66$ rather than 1,
so $1-\text{pass@}8$ is not the fraction of problems that all eight samples
missed. The separate evaluation reports pass@16 and maj@16 directly from the 16
samples. A selection-corrected peak subtracts from a run's maximum the expected
maximum of $m$ draws from a flat curve with that run's own noise,
$\hat\sigma\,\mathbb{E}[\max_{i\le m} z_i]$ with $z_i\sim\mathcal{N}(0,1)$, $m$
the run's number of evaluations and $\hat\sigma$ the standard deviation of its
successive differences divided by $\sqrt2$.

\paragraph{Late-training averages.}
Peaks select the best of about 31 noisy evaluations. Table~\ref{tab:last10} gives
the complementary estimator, the mean of each run's last ten evaluations, which
removes the selection but mixes in any late drift. Every arm sits 2.5 to 7.3
points below its peak, most on AIME 2026 pass@8. On these averages the
zero-label runs are level with supervised PPO on AIME 2026 pass@1 (22.1 against
22.5) and ahead on pass@8 (37.2 against 34.4), while supervised PPO keeps a
2.4-point edge on AMC 2023 pass@1. The ten-evaluation windows are not aligned
across runs; restricting every run to steps 210--300 changes no entry by more than
0.9 points.

\begin{table}[h]
\centering
\footnotesize
\setlength{\tabcolsep}{5pt}
\caption{\textbf{Late-training validation.} Accuracy (\%) averaged over each
run's last ten evaluations, the complement to the peaks of
Table~\ref{tab:dynamics}. All rows train under the $5{,}120$-token cap except the
last two, the $4{,}096$-token-cap pair added later (Appendix~\ref{app:budget}).}
\label{tab:last10}
\begin{tabular}{lrcccc}
\toprule
\rowcolor{hdrbg}
& & \multicolumn{2}{c}{AIME 2026} & \multicolumn{2}{c}{AMC 2023} \\
\rowcolor{hdrbg}
Arm & Window & pass@1 & pass@8 & pass@1 & pass@8 \\
\midrule
Initial policy $\pi_{\theta_0}$ & step 0 & 21.2 & 32.1 & 73.6 & 89.5 \\
\midrule
PPO, 100\% labels & 210--300 & 22.5 & 34.4 & \textbf{74.7} & 90.4 \\
\car, 50\% labels & 240--330 & \textbf{23.1} & \textbf{39.1} & 74.5 & \textbf{90.9} \\
\midrule
\car, 0\% labels, run 1 & 220--310 & 22.6 & 37.3 & 73.3 & 89.9 \\
\car, 0\% labels, run 2 & 230--320 & 21.8 & 37.0 & 71.2 & 89.2 \\
\car, 0\% labels, run 3 & 280--370 & 21.9 & 37.4 & 72.3 & 90.0 \\
\rowcolor{gray!12}
\textbf{Reward-free} {\scriptsize(mean $\pm$ s.d.)} & & 22.1\,{\scriptsize$\pm$0.4} & 37.2\,{\scriptsize$\pm$0.2} & 72.3\,{\scriptsize$\pm$1.1} & 89.7\,{\scriptsize$\pm$0.4} \\
\midrule
PPO, 100\% labels, $4{,}096$ cap & 220--310 & 21.4 & 33.9 & 72.6 & 89.4 \\
\car, 0\% labels, $4{,}096$ cap & 220--310 & 22.2 & 38.1 & 73.6 & 89.1 \\
\bottomrule
\end{tabular}
\end{table}

\paragraph{Extended metrics at step 300.}
Table~\ref{tab:extended} adds majority voting, pass@16 and generation length to
the separate evaluation of Table~\ref{tab:main}. The three reward regimes stay
close under maj@16. The largest gap is AIME 2026 pass@16, 50.0 for supervised PPO
against 40.0 for the zero-label checkpoint, which is three problems out of 30.
All RL checkpoints keep the short, rarely truncated generation of the initial
policy. The zero-label checkpoint writes somewhat longer answers, with median
length 13--19\% above supervised PPO on AIME and GPQA and 3\% below on AMC, and
truncates at most 2.2\% of them.

\begin{table}[h]
\centering
\footnotesize
\setlength{\tabcolsep}{4pt}
\caption{\textbf{Extended metrics of the step-300 checkpoints.} Same
evaluation as Table~\ref{tab:main}: 16 samples, temperature 0.7,
$12{,}288$-token budget.}
\label{tab:extended}
\begin{tabular}{lcccc|cc}
\toprule
\rowcolor{hdrbg}
& \multicolumn{4}{c|}{maj@16 / pass@16 (\%)} & \multicolumn{2}{c}{median tokens / truncated (\%)} \\
\rowcolor{hdrbg}
Model & AIME25 & AIME26 & AMC23 & GPQA & AIME26 & GPQA \\
\midrule
+ SFT & 20.0 / 50.0 & 20.0 / 43.3 & 60.0 / 92.5 & 26.3 / 89.4 & 12{,}288 / 63.5 & 10{,}197 / 40.7 \\
Initial policy $\pi_{\theta_0}$ & 20.0 / 40.0 & 20.0 / 33.3 & 72.5 / 92.5 & 37.9 / 87.9 & 3{,}963 / 4.0 & 2{,}900 / 2.1 \\
\midrule
PPO, 100\% labels & 23.3 / 50.0 & 23.3 / 50.0 & 70.0 / 95.0 & 40.4 / 86.9 & 4{,}151 / 0.8 & 2{,}923 / 0.4 \\
\rowcolor{carbg}
\car, 50\% labels & 20.0 / 46.7 & 20.0 / 43.3 & 75.0 / 95.0 & 43.9 / 89.9 & 4{,}586 / 1.0 & 2{,}909 / 1.0 \\
\rowcolor{carbg}
\car, 0\% labels & 23.3 / 46.7 & 20.0 / 40.0 & 70.0 / 90.0 & 41.9 / 89.9 & 4{,}920 / 1.2 & 3{,}359 / 1.3 \\
\bottomrule
\end{tabular}
\end{table}

\section{Training dynamics}
\label{app:dynamics}

\begin{figure}[h]
\centering
\includegraphics[width=\textwidth]{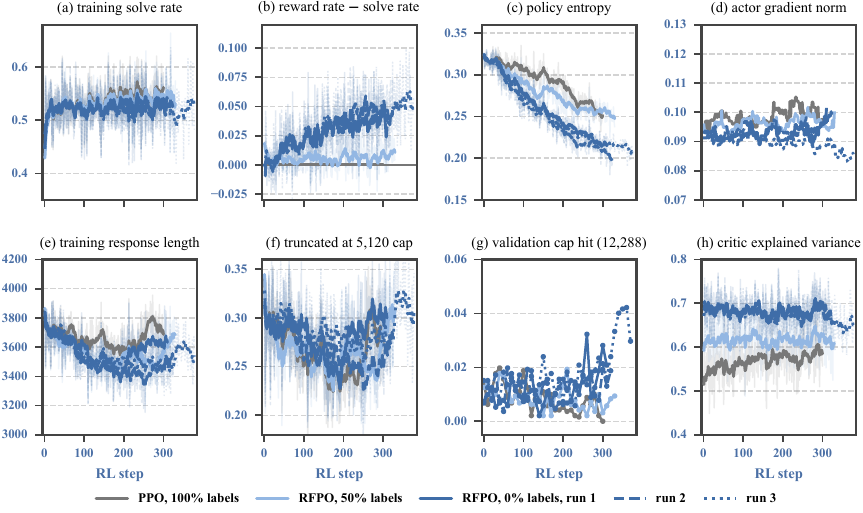}
\caption{\textbf{Training dynamics of the main arms.} Smoothed curves over faint
raw traces; (d) is a 21-step running median and (g) is raw. (b) is the positive
rate of the reward actually used minus the verifier solve rate on the same
rollouts, and is zero by construction for supervised PPO. In (h) the supervised
critic is scored against verifier returns and the frozen critic against the
returns of its own reward, so the two levels are not comparable.}
\label{fig:internals}
\end{figure}

Figure~\ref{fig:internals} follows eight training-time quantities for supervised
PPO, the half-labeled arm and the three zero-label runs; the calibration
ablations have their own dynamics in Appendix~\ref{app:reward}.

\paragraph{Optimization.}
Nothing in the optimizer statistics separates the arms. The actor gradient norm
holds near 0.09--0.10 throughout. Entropy declines gradually in every arm, faster
under \car, from 0.32 to 0.21--0.22 by step 300 against 0.25 for supervised PPO
and 0.26 for the half-labeled arm. Training response length is flat under
supervised PPO and drifts 2--7\% shorter in the zero-label runs, while 26--30\% of
training rollouts stay truncated at the $5{,}120$-token cap. At validation, fewer
than 2\% of responses hit the $12{,}288$-token budget through step 300; only the
longest run rises to about 4\% over its final 60 steps.

\paragraph{The frozen reward as the policy moves.}
Two panels show what freezing costs, and both are small. Supervised PPO raises
its training solve rate from 52.2\% to 55.8\% over 300 steps, the zero-label runs
from about 52\% to 53--54\%: \car optimizes a proxy, and it moves the verifier's
number on training prompts less than the verifier itself does, even though
held-out accuracy ends level (Table~\ref{tab:last10}). The proxy also loosens.
The positive rate of the binary reward starts 0.5--1.0 points above the verifier
solve rate, as calibration intends, and climbs to 4.5--5.4 points above it by
step 300: a threshold fitted on $\pi_{\theta_0}$ admits more false positives as
the policy moves toward rollouts the critic scores highly. The critic's ranking
meanwhile holds (Appendix~\ref{app:drift}), so what drifts is the cut, not the
ordering. In the half-labeled arm the gap ends at 1.3 points, or 2.6 on the
critic-scored half alone, about half the zero-label value; this is the defense
that partial labels provide in \S\ref{sec:results}.

\section{The critic: measurements and robustness}
\label{app:critic}

This appendix defines the critic metrics used in \S\ref{sec:identity}, shows the
critic's training diagnostics, and gives the full tables behind the offline checks,
the drift test and the generation-budget test.

\paragraph{Metrics.}
Each rollout $i$ has a critic score $v_i$, a verifier label $y_i\in\{0,1\}$ and a
problem index $g_i$. \emph{Overall AUC} is the Mann--Whitney probability that a
correct rollout scores above an incorrect one, pooled over all problems; it
credits a critic that only recognizes problem difficulty. \emph{Within-problem
AUC} computes the same statistic inside each problem with both outcomes and at
least three rollouts, and averages over those problems; it asks the question a
reward must answer, which of several attempts at the same problem succeeded. The
\emph{truncated-only} and \emph{finished-only} variants restrict each problem to
rollouts that did or did not reach $0.99$ of the generation cap. \emph{Explained
variance} is $1-\mathrm{Var}(y-v)/\mathrm{Var}(y)$. Unlike the two AUCs it depends
on the scale of $v$, not only on its order. $\tau^\ast$ is the threshold on the
raw score that maximizes balanced accuracy, and \emph{precision} is the fraction
of rollouts in fact correct among those whose raw score exceeds $\tau=0.5215$.

\subsection{Critic training diagnostics}
\begin{figure}[h]
\centering
\includegraphics[width=\textwidth]{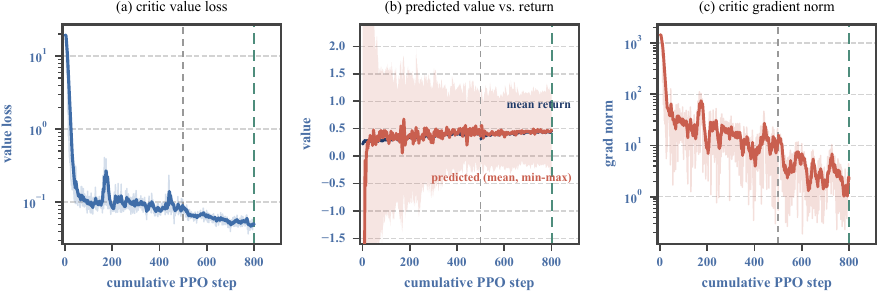}
\caption{\textbf{Critic diagnostics across both pretraining phases.} The gray
dashed line marks the phase boundary at step 500 and the green line the frozen
critic at step 800. In (b) the early predictions below $-1.6$ are off the axis.}
\label{fig:criticinternals}
\end{figure}

Figure~\ref{fig:criticinternals} complements Figure~\ref{fig:provenance}. Value
loss falls from 19 to about 0.05, the mean prediction converges from $-5.5$ onto
the mean return within about 20 steps, and the critic gradient norm decays by
three orders of magnitude. The spread of predictions keeps narrowing long after
the mean has converged, which is why a critic at step 200 already ranks rollouts
but does not yet produce values that can be read as probabilities
(Figure~\ref{fig:trust}a).

\subsection{The measurement plane}
\label{app:plane}
The checkpoint analysis of \S\ref{sec:identity} crosses nine critic checkpoints
(cumulative steps 100--900) with ten actor checkpoints (the same nine steps plus a
second step-600 actor from a parallel branch). To keep the two axes from
confounding each other, the prompts are held fixed rather than the rollouts: each
actor regenerates the same 512 problems with 32 samples at a $5{,}120$-token cap,
giving $16{,}384$ rollouts and 323--360 problems with both outcomes per cell, and
every critic scores every cell. Prompt-clustered bootstrap intervals on
within-problem AUC have half-widths of 0.017--0.026 per cell.
Table~\ref{tab:plane} averages each critic over the ten actors.

\begin{table}[h]
\centering
\footnotesize
\setlength{\tabcolsep}{4pt}
\caption{\textbf{Measurement plane, per critic.} Each row averages ten actor
distributions of $16{,}384$ rollouts. The shaded row is the frozen critic.
Truncated rollouts are 19--54\% of a cell depending on the actor.}
\label{tab:plane}
\begin{tabular}{lcccccccc}
\toprule
\rowcolor{hdrbg}
& & \multicolumn{3}{c}{Within-problem AUC} & & & & \\
\rowcolor{hdrbg}
Critic step & AUC & all & truncated & finished & EV & $\tau^\ast$ & Precision & $v\notin[0,1]$ \\
\midrule
100 & 0.913 & 0.765 & 0.800 & 0.608 & 0.467 & 0.956 & 0.532 & 0.443 \\
200 & 0.923 & 0.776 & 0.794 & 0.640 & 0.460 & 0.462 & 0.796 & 0.044 \\
300 & 0.930 & 0.789 & 0.802 & 0.667 & 0.521 & 1.082 & 0.540 & 0.556 \\
400 & 0.933 & 0.797 & 0.820 & 0.678 & 0.568 & 0.809 & 0.683 & 0.325 \\
500 & 0.937 & 0.796 & 0.808 & 0.683 & 0.593 & 0.757 & 0.673 & 0.295 \\
600 & 0.935 & 0.793 & 0.808 & 0.675 & 0.537 & 0.469 & 0.819 & 0.094 \\
700 & 0.956 & 0.824 & 0.830 & 0.729 & 0.688 & 0.554 & 0.859 & 0.202 \\
\rowcolor{carbg}
800 & 0.960 & 0.837 & 0.832 & 0.746 & 0.702 & 0.540 & 0.875 & 0.077 \\
900 & 0.965 & 0.842 & 0.835 & 0.757 & 0.721 & 0.699 & 0.862 & 0.271 \\
\bottomrule
\end{tabular}
\end{table}

Three things in the table matter beyond the main text. First, ranking and
calibration come apart: from step 200 to step 300 overall AUC rises while
$\tau^\ast$ jumps from 0.46 to 1.08 and precision at the deployed threshold falls
from 0.80 to 0.54, because more than half of the step-300 scores leave $[0,1]$.
This is why the threshold must be fitted on the deployed critic
(\S\ref{sec:calib}). Second, within-problem AUC on finished rollouts is lower than
on all rollouts, 0.61 to 0.76. Truncated rollouts are rarely correct (8--14\%),
so a problem that mixes finished and truncated attempts offers easy contrasts,
and removing the truncated attempts removes them. The finished-only ranking rises
with critic training as steeply as the others do, and the truncated-only column
shows that the critic ranks attempts that never finished just as well as the
full set, so the signal is not truncation detection. Third, explained variance
tracks within-problem AUC across the nine checkpoints ($r{=}0.98$ offline, $0.91$
for the explained variance logged during training), the basis for using it as a
training-time indicator.

\subsection{Prefix scan}
\label{app:prefix}
Table~\ref{tab:prefixfull} gives the prefix scan of Figure~\ref{fig:trust}e: the
frozen critic scores prefixes of $1{,}024$ rollouts from $\pi_{\theta_0}$ (128
problems, 8 responses each, 46 problems with both outcomes) against the final
correctness of the full rollout. Up to about $2{,}000$ tokens nearly all variance in
the score is between problems: the critic recognizes how hard the problem is
before it can tell attempts apart. Within-problem ranking then rises with the
visible reasoning, and on prefixes that have not yet produced an answer it
reaches 0.84--0.92 from $4{,}096$ tokens on, on a shrinking number of problems.

\begin{table}[h]
\centering
\footnotesize
\setlength{\tabcolsep}{5pt}
\caption{\textbf{Prefix scan.} ``No answer yet'' restricts to rollouts still
running at that prefix length; ``between-problem share'' is the fraction of score
variance explained by problem identity; ``finished'' is the fraction of rollouts
already complete.}
\label{tab:prefixfull}
\begin{tabular}{rccccc}
\toprule
\rowcolor{hdrbg}
Prefix (tokens) & AUC & Within & Within, no answer yet (problems) & Between-problem share & Finished \\
\midrule
256 & 0.958 & 0.526 & 0.521 (46) & 0.996 & 0.003 \\
512 & 0.961 & 0.477 & 0.470 (46) & 0.994 & 0.003 \\
1{,}024 & 0.967 & 0.600 & 0.603 (46) & 0.982 & 0.004 \\
2{,}048 & 0.967 & 0.647 & 0.645 (46) & 0.955 & 0.141 \\
3{,}072 & 0.969 & 0.735 & 0.700 (41) & 0.924 & 0.395 \\
4{,}096 & 0.976 & 0.836 & 0.844 (30) & 0.905 & 0.615 \\
4{,}608 & 0.978 & 0.852 & 0.909 (21) & 0.903 & 0.686 \\
4{,}864 & 0.979 & 0.846 & 0.921 (14) & 0.904 & 0.721 \\
5{,}120 & 0.979 & 0.843 & 0.885 (12) & 0.904 & 0.762 \\
\bottomrule
\end{tabular}
\end{table}

\subsection{Frozen-critic reliability across policy drift}
\label{app:drift}
\begin{figure}[h]
\centering
\includegraphics[width=\textwidth]{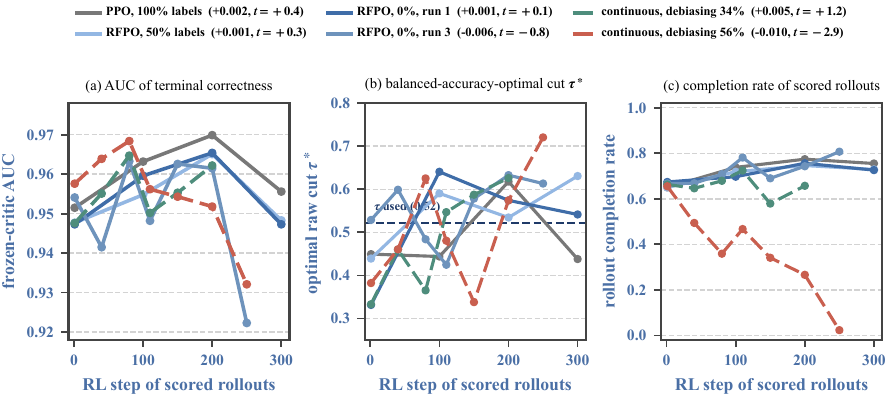}
\caption{\textbf{Frozen critic on each arm's own rollouts.} Legend gives the AUC
slope per 100 steps and its $t$ statistic. The dashed line in (b) is the deployed
$\tau=0.5215$. The two continuous-score arms are those of
Appendix~\ref{app:reward}.}
\label{fig:drift}
\end{figure}

We re-apply the frozen critic to each arm's own training rollouts at a grid of
steps, $1{,}024$ rollouts per point (Figure~\ref{fig:drift}). One AUC on $1{,}024$
rollouts has a standard error of about 0.009, so we test for a trend rather than
reading a range. Per 100 steps the AUC slope is $+0.002$ ($t{=}0.44$) for
supervised PPO, $+0.001$ ($t{=}0.28$) for the half-labeled arm, and $+0.001$
($t{=}0.11$) and $-0.006$ ($t{=}{-}0.79$) for the two zero-label runs scored; none
is significant. The continuous-score arm with debiasing strength 34\% is also flat
($+0.005$, $t{=}1.20$). The one significant decline is the 56\% arm ($-0.010$,
$t{=}{-}2.89$, from 0.968 at step 80 to 0.932 at step 250), whose completion rate
falls from 0.36 to 0.02 over the same span: the critic is handed a distribution of
unfinished responses it was never fitted on. Ranking is stable wherever the policy
stays healthy. The optimal cut is not: in the healthy arms $\tau^\ast$ moves
between 0.33 and 0.64 while AUC stays flat, which is the calibration drift that
Appendix~\ref{app:dynamics} sees as a rising reward rate.

\subsection{The generation-budget axis}
\label{app:budget}
\begin{figure}[h]
\centering
\includegraphics[width=\textwidth]{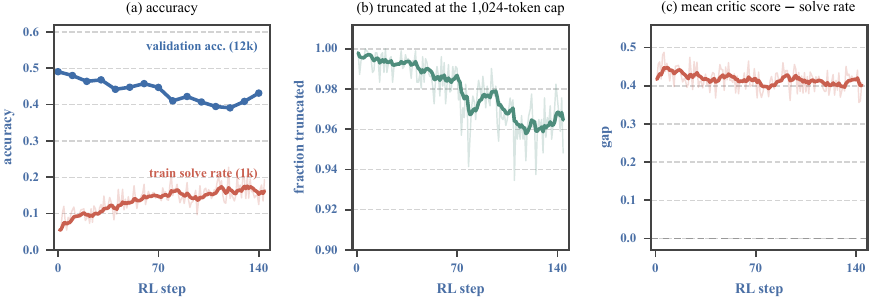}
\caption{\textbf{Training under a $1{,}024$-token cap.} (a) Validation at
$12{,}288$ tokens (macro average) against training solve rate; (b) fraction of
training rollouts truncated; (c) mean critic score minus verifier solve rate.}
\label{fig:shortrollout}
\end{figure}

Table~\ref{tab:budget-axis} compares really generating under a cap with the
cheaper proxy of truncating long rollouts to the same length, both scored by the
frozen critic on the step-700 actor. From $5{,}120$ tokens up the two agree, and
the critic's precision is flat at 0.89--0.90 while the finished fraction grows
from 0.74 to 0.97. At $1{,}024$ tokens they point in opposite directions: under a
real cap only 0.7\% of rollouts finish, explained variance is $-0.49$ and precision
0.21, while the proxy reads $+0.55$ and 0.88. The proxy would have cleared a
setting in which the critic is not usable. A training run under the real
$1{,}024$-token cap confirms it (Figure~\ref{fig:shortrollout}): the training solve
rate climbs from 5.5\% to 19.3\% while validation at the full budget falls from
49.1\% to 39.1\%, and the mean critic score sits about 0.4 above the solve rate
from the first step.

\begin{table}[h]
\centering
\footnotesize
\setlength{\tabcolsep}{5pt}
\caption{\textbf{Real caps versus truncation.} $16{,}384$ rollouts per column;
the proxy truncates $12{,}288$-token rollouts to the cap.}
\label{tab:budget-axis}
\begin{tabular}{lccccc}
\toprule
\rowcolor{hdrbg}
Generation cap (tokens) & 1{,}024 & 2{,}048 & 5{,}120 & 8{,}192 & 12{,}288 \\
\midrule
Finished, real cap & 0.007 & 0.179 & 0.735 & 0.941 & 0.967 \\
EV, real cap & $-0.492$ & $+0.496$ & $+0.701$ & $+0.669$ & $+0.663$ \\
EV, truncation proxy & $+0.554$ & $+0.592$ & $+0.649$ & $+0.664$ & --- \\
Precision, real cap & 0.213 & 0.675 & 0.895 & 0.896 & 0.894 \\
Precision, truncation proxy & 0.879 & 0.919 & 0.905 & 0.897 & --- \\
\bottomrule
\end{tabular}
\end{table}

\paragraph{Training under a $4{,}096$-token cap.}
The $1{,}024$-token run fails because almost nothing it generates finishes. A
$4{,}096$-token cap tests the regime the method is meant for, in which a large
share of every batch is unfinished but the rollouts still resemble those the
critic was fitted on. We keep the initial pair and every setting of
\S\ref{sec:setup} except the training cap, and refit $b(\ell)$ and $\tau$ on 512
step-1 rollouts at that cap. The deployed $5{,}120$-token fit leaves short and
long responses of the same outcome scoring alike (median split within each
correctness class, difference $0.000$); refitted at $4{,}096$ tokens on six length
bins, the offsets leave a difference of 0.049, and scaling them by 1.7 restores
it to zero ($-0.000$; 0.028 at 1.3, $-0.021$ at 2.0). We train one zero-label run
with that correction ($\tau=0.6526$; the binary rule agrees with the verifier on
90.6\% of the calibration sample) and supervised PPO under the same cap, both for
310 steps, covering the 300-step comparison window of \S\ref{sec:setup}
(Figure~\ref{fig:cap4096} in \S\ref{sec:results}, Table~\ref{tab:cap4096}).
At step~1, 50\% of the run's training rollouts are truncated, against 31--33\% at
$5{,}120$ tokens, and the run keeps them there: responses hold their length
(3{,}313 tokens over the first ten steps, 3{,}462 over the last hundred), and
48\% of its training rollouts over the window are unfinished (50\% over the last
hundred steps). The reward stays calibrated all the same: the mean reward exceeds
the verifier solve rate by 0.014 on average over the window, against
0.027--0.030 for the $5{,}120$-token runs and about 0.4 for the $1{,}024$-token
run. Macro validation averages 47.6\% over steps 10--300, level with the
$5{,}120$-token runs (47.2--47.4\%) and above supervised PPO under the same cap
(46.6\%).
On the same node its median step takes 395\,s against 490\,s for supervised
PPO under the same cap (264 against 327 GPU-hours for 300 steps): generation is
slower because its responses stay longer (162 against 135\,s), and the 134\,s
critic update that PPO needs is gone. Supervised PPO with the full
$5{,}120$-token cap averages 48.3\% at 388 GPU-hours (Appendix~\ref{app:cost}).

\section{Calibration ablations: turning the critic's score into a reward}
\label{app:reward}\label{sec:reward}

The frozen critic returns one number per rollout, and something has to decide
what reward the policy receives for it. The obstacle is a length bias, and it
has to be separated from something that is not a bias. On the $16{,}384$
rollouts from the deployed pair, the mean score falls from 0.88 in the shortest
length decile to 0.13 at the cap, and accuracy falls with it, from 0.92 to 0.14:
length and correctness are mixed, and much of the score's length trend is
correctness. Not all of it. With the outcome held fixed the score still falls
with length, among correct responses from 0.92 to 0.61 before the cap and among
incorrect ones from 0.37 to 0.17; among responses to the same problem with the
same outcome, score and length correlate at $-0.43$. That residual trend is the
bias.

A bias is harmless to read and unsafe to optimize against. Length is something
the policy controls directly, and a shorter answer is not a more correct one; a
reward that carries the bias teaches the policy to move length instead of
correctness. A policy rewarded with the raw score does exactly that: it stops
reasoning, shortening responses by 73\% within 100 steps, while the reward it
collects drifts from $+0.19$ to $+0.70$ above the batch's true accuracy and
validation falls 17 points by step~100 and 32 by step~230.

Eq.~(\ref{eq:car}) closes that channel twice. Subtracting $b(\ell)$, the
within-class offsets of Table~\ref{tab:lenbaseline}, removes the length trend
inside each correctness class and leaves the gap between classes in place.
Because the two trends overlap in the pooled data, the separation costs a little
ranking: AUC falls from 0.959 to 0.941. It is also not exact. After it, score and
length correlate at $-0.26$ among correct responses (from $-0.58$) and at
$+0.22$ among incorrect ones (from $-0.32$), so some length dependence remains,
pointing different ways in the two classes. The indicator then caps what that
remainder can earn. Both pieces are load-bearing, and the rest of this section
removes them one at a time (Appendix~\ref{app:ladder}).

\begin{figure}[t]
\centering
\includegraphics[width=\textwidth]{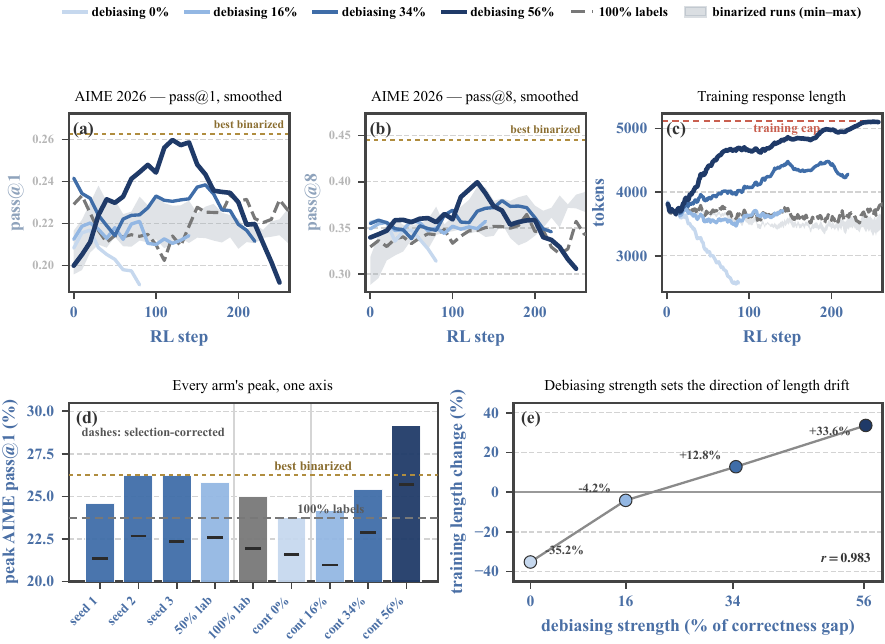}
\caption{\textbf{Removing the threshold raises the ceiling and loses the
policy.} Four arms use the debiased score directly as the reward ($\hat r = \vB -
b(\ell)$, no threshold, no indicator) and differ only in how much of the
debiasing curve they apply. The resulting \emph{debiasing strength}---the
within-group spread of the debiasing term as a fraction of the score gap between
correct and incorrect responses---is measured for each arm rather than set. Length
and accuracy both change across these arms; the figure does not attribute one to
the other. Gray band: the three binarized zero-label runs; dashed gray:
PPO on 100\% verifier labels; gold line: the best value any binarized run
reached. (a--b) Held-out accuracy, exponentially smoothed; the fully debiased arm
leads the sweep on both metrics through most of its run. Peaks are single
evaluations and cannot be read off a smoothed curve, so they are reported in (d)
instead. (c) The fully debiased arm lengthens responses by 34\% into the training cap,
while the arm with no debiasing shortens them by 35\%. (d) Every arm's peak on one
axis with its selection-corrected value marked; the fully debiased arm reaches 29.2,
above the 25.0 of fully supervised PPO. (e) Debiasing strength against the change
in training response length.}
\label{fig:contv}
\end{figure}

\paragraph{Remove the indicator and the leftover length dependence sets the drift.}
Figure~\ref{fig:contv} drops the indicator and varies how much of the debiasing
curve the reward applies. These arms use a curve refitted on the $12{,}046$
finished rollouts among the $16{,}384$, running from $+0.108$ at the shortest
decile to $-0.129$ at the longest. One arm applies no debiasing. Two apply the
curve only up to $2{,}828$ and $3{,}558$ tokens and hold it constant beyond. One
applies all of it.

How strongly each choice debiases the reward is then \emph{measured}, not set.
Within a group of eight responses we compare the spread of the debiasing term
against the score gap between correct and incorrect responses (0.256 in these
rollouts), and call the ratio the \emph{debiasing strength}. The four arms come
out at 0\%, 16\%, 34\% and 56\%.

Under a continuous reward the policy climbs whatever length dependence the reward
still has, and the direction follows the debiasing strength. Response length goes
$-35.2\%$, $-4.2\%$, $+12.8\%$ and $+33.6\%$ ($r{=}0.983$ with strength, measured
on every step): left undebiased, the reward favors short responses and the policy
shortens; fully debiased, the policy lengthens. Neither direction says that
shorter or longer answers are more correct; both are the policy exploiting the
reward. Held-out accuracy peaks in every arm, and none keeps what it reaches:

\begin{center}
\small
\begin{tabular}{lccc}
\toprule
\rowcolor{hdrbg}
Debiasing strength & Peak AIME pass@1 & Final AIME pass@1 & Length change \\
\midrule
0\%  & 23.8 & 17.1 & $-35.2\%$ \\
16\% & 24.2 & 22.1 & $-4.2\%$ \\
34\% & 25.4 & 19.6 & $+12.8\%$ \\
56\% & \textbf{29.2} & 16.2 & $+33.6\%$ \\
\bottomrule
\end{tabular}
\end{center}

\noindent For reference, the shared initial policy scores 21.2 on this suite and
PPO trained on every label peaks at 25.0. Three of the four arms end below where
they started.

\noindent Length and accuracy move together here because the debiasing strength
moves both, so the table does not say that longer answers are better; in the
data the critic was fitted on, long responses are less often correct. What the
table does show is that no setting of the continuous reward both
gains substantially and keeps what it gains, and that the two ends fail in
opposite directions---one shortens and loses, the other lengthens and loses.

The fully debiased arm reaches \textbf{29.2} on AIME~2026 pass@1, the highest
AIME 2026 score anywhere in this paper and above the 25.0 of PPO trained on
every label. That is not an artifact of having more checkpoints to choose from:
it has fewer evaluations than any binarized arm, and correcting each arm for the
expected maximum of its own number of draws leaves the ordering unchanged (25.7
against 21.4--22.7 binarized). It also ends at 16.2, below the initial
policy's 21.2.

What the sweep does isolate is the indicator's job. The deployed rule debiases
more strongly than any arm above (91\%) and still holds response length within
13\% of its starting value for 310 steps, because above the cut extra score earns
nothing. Under a continuous reward the debiasing strength sets which way the
policy drifts; the indicator decides whether it drifts at all.

\paragraph{The decline is detectable, but not by the obvious signals.}
Held-out accuracy falls, so any validated run sees it; what is hidden is the
cause. The model is neither producing garbage nor fooling the answer checker:
verbatim repetition stays at or below 1\% of 60-character shingles, restricting
the answer search to the final boxed expression moves train-batch accuracy by at
most 5.1 points with no trend, the mean raw score stays within 0.075 of the
verifier solve rate
for all 257 steps, and entropy is 0.322 at the peak against 0.314 at the end.
What does track the damage is answer restatement: the number of times a response
asserts its own answer rises from 1.8 to 67.9 and correlates with held-out
accuracy at $r{=}{-}0.87$, where length manages $-0.51$
(Appendix~\ref{app:collapse}).

\paragraph{What this leaves.}
Without the indicator the reward does work---the same frozen critic, used
directly as the reward, reaches an accuracy no labeled run in this paper reaches---but it
works in a way we can neither take apart nor hold onto. We cannot take it apart
because the one setting that buys the gain also moves response length, so what
the policy actually learned is not identifiable from these runs. We cannot hold
onto it because every arm declines afterwards, and the arm that gains the most
declines the most. Binarizing gives up that ceiling and takes back both problems
at once: the reward becomes a decision rather than a quantity, the length
channel closes, and three runs train for over 300 steps without drifting.
Recovering the ceiling safely would need a validated stopping rule.
Restatement---which tracks the damage closely and rises well before an arm gives
up its gain---is the natural quantity to build one from, and we leave that to
future work.

\subsection{Removing the calibration step by step}
\label{app:ladder}
\begin{figure}[h]
\centering
\includegraphics[width=\textwidth]{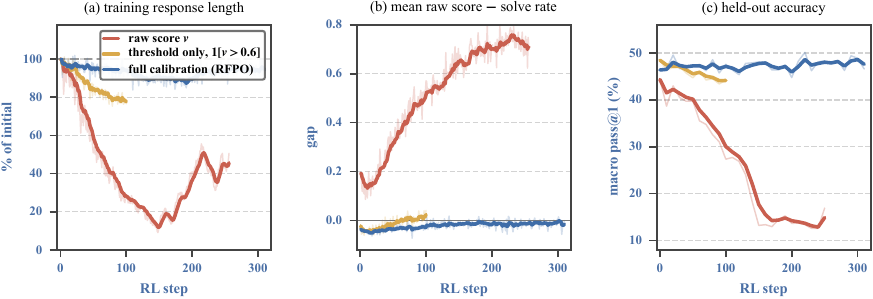}
\caption{\textbf{The calibration ladder.} Raw score as reward, a fixed threshold
without debiasing, and the full rule of Eq.~(\ref{eq:car}); (b) is the mean
raw terminal score minus the verifier solve rate on the training batch. The
raw-score run used an earlier actor--critic pair and an $8{,}192$-token training
cap, and starts from a lower validation point.}
\label{fig:ladder}
\end{figure}

Figure~\ref{fig:ladder} removes the two pieces of Eq.~(\ref{eq:car}) in turn. With
the raw score as the reward, response length falls 73\% by step 100, the mean
score climbs from $+0.19$ to $+0.70$ above the verifier solve rate, and held-out
accuracy falls 17 points by step 100 and 32 by step 230. A fixed threshold with no
debiasing, $\mathbf{1}[v>0.6]$, keeps the score honest on average but not the
length: responses shorten by 20\% in 100 steps and accuracy drifts down 4 points.
Only with both pieces does length hold and accuracy hold. The raw-score run is the
oldest in the paper: it used an earlier actor--critic pair and an $8{,}192$-token
cap. The continuous arm with no debiasing
in Table~\ref{tab:contv} reproduces its failure on the deployed pair and
protocol, shortening responses by 35\% in 87 steps.

\subsection{Continuous-score arms and a group-relative variant}
\label{app:grpo}
Table~\ref{tab:contv} lists the continuous arms. They differ from the zero-label
runs only in dropping the indicator; actor, critic, data order, cap and optimizer
settings are shared. A fifth arm replaces GAE with group-relative advantages at
the 56\% setting. It reaches a similar peak (28.3 on AIME pass@1) sooner, with the
best macro average at step 40 after 5.5 hours against step 110 and 16.3 hours for
the GAE arm, and it also runs into the cap sooner: more than 90\% of its
rollouts are truncated from step 150, against step 231 with GAE. Because $b(\ell)$
depends on each response's own length, the debiasing term survives group-mean
centering, and dividing by the group standard deviation enlarges its share of
the advantage. The variant also gives up the identity $A_t=\hat r-\vB(x,y_{\le t})$ that
lets one frozen network serve as both reward and baseline, so we do not adopt it.

\begin{table}[h]
\centering
\footnotesize
\setlength{\tabcolsep}{4.5pt}
\caption{\textbf{Continuous-score arms.} Accuracies are validation pass@1 in
points; length change compares mean training response length at the last and
first step;
``truncated'' is the fraction of training rollouts at the $5{,}120$ cap over the
last five steps; ``up to'' means the debiasing term is held constant beyond that
length. The arms start at 20.0--24.2 on AIME and 46.7--48.0 macro.}
\label{tab:contv}
\begin{tabular}{lcrccccl}
\toprule
\rowcolor{hdrbg}
Arm & Strength & Steps & AIME peak / final & Macro peak & $\Delta$ length & Truncated \\
\midrule
No debiasing & 0\% & 87 & 23.8 / 17.1 & 48.8 & $-35.2\%$ & 0.05 \\
Debiased up to $2{,}828$ & 16\% & 141 & 24.2 / 22.1 & 49.3 & $-4.2\%$ & 0.23 \\
Debiased up to $3{,}558$ & 34\% & 220 & 25.4 / 19.6 & 51.1 & $+12.8\%$ & 0.26 \\
\rowcolor{carbg}
Full debiasing curve & 56\% & 257 & \textbf{29.2} / 16.2 & \textbf{53.3} & $+33.6\%$ & 0.97 \\
56\%, group-relative & 56\% & 161 & 28.3 / 15.8 & 52.4 & $+33.4\%$ & 0.97 \\
\bottomrule
\end{tabular}
\end{table}

\subsection{Anatomy of the decline}
\label{app:collapse}
\begin{figure}[h]
\centering
\includegraphics[width=\textwidth]{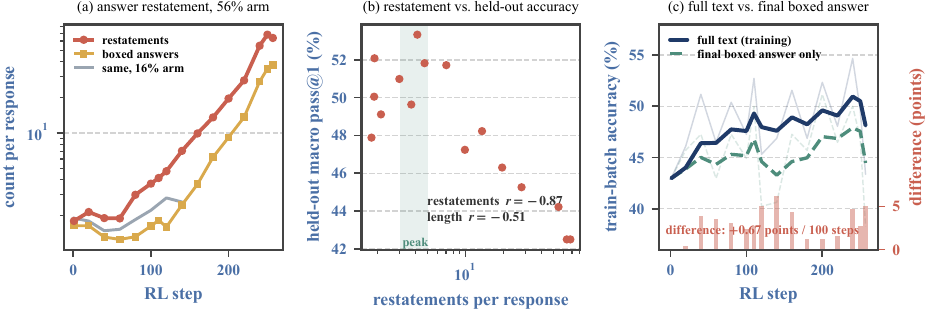}
\caption{\textbf{Restatement during the decline.} (a) Restatements and boxed
answers per response in the 56\% arm, with restatements in the 16\% arm for
comparison. (b) Restatement against held-out macro accuracy across checkpoints.
(c) Train-batch accuracy scored on the full text and on the final boxed answer
only; bars are their difference.}
\label{fig:collapse}
\end{figure}

Table~\ref{tab:collapse} and Figure~\ref{fig:collapse} inspect 256 training
rollouts per checkpoint of the 56\% arm. What changes is answer restatement: the
number of times a response asserts its answer rises from 1.8 to 68, boxed answers
from 1.6 to 35, and the share of responses that never close their reasoning block
from 25\% to 91\%, while verbatim repetition stays at 1\% of 60-character
shingles. Across checkpoints restatement tracks held-out accuracy at $r{=}{-}0.87$,
against $-0.51$ for length. The verifier is not being gamed: scoring only the
final boxed answer changes train-batch accuracy by at most 5.1 points, with no
trend. The same measurements on the 16\% arm stay flat over its 141 steps
(restatements 1.9 to 2.6, boxed answers 1.7 to 2.2, unclosed reasoning 20\% to
15\%), so the behavior belongs to the heavy-weight arm, not to continuous rewards
in general.

\begin{table}[h]
\centering
\footnotesize
\setlength{\tabcolsep}{5pt}
\caption{\textbf{What the 56\% arm's decline is made of.} 256 inspected rollouts
per checkpoint; length, score gap and entropy are from the training log. Neither
entropy nor the score gap registers the change.}
\label{tab:collapse}
\begin{tabular}{lccc}
\toprule
\rowcolor{hdrbg}
56\% arm & Step 1 & Peak (step 110) & Step 250 \\
\midrule
Restatements per response & 1.79 & 4.14 & \textbf{67.94} \\
Boxed answers per response & 1.63 & 1.82 & 34.92 \\
Never closes its reasoning block & 0.250 & 0.363 & \textbf{0.914} \\
Verbatim repetition (60-character shingles) & 0.000 & 0.000 & 0.010 \\
Mean response length (tokens) & 3{,}812 & 4{,}556 & 5{,}107 \\
Train-batch accuracy, full text & 0.430 & 0.527 & 0.496 \\
Train-batch accuracy, final boxed answer only & 0.430 & 0.500 & 0.469 \\
Mean raw score $-$ solve rate & $-0.037$ & $-0.060$ & $+0.049$ \\
Policy entropy & 0.322 & 0.322 & 0.312 \\
\bottomrule
\end{tabular}
\end{table}

\section{Computation and memory cost}
\label{app:cost}

\paragraph{Accounting.}
With $P$ parameters shared by policy and critic and $T$ tokens in a batch, a
forward pass costs about $2PT$ and a forward--backward about $6PT$. Beyond
generation, which is identical in both arms, a supervised PPO step runs the
old-log-probability forward ($2PT$), the value forward ($2PT$), the actor update
($6PT$) and the critic update ($6PT$), $16PT$ in total. \car keeps the first three
and drops the fourth, $10PT$, a 37\% reduction; with activation recomputation each
trained module costs $8PT$ and the reduction is 40\%. The reward needs no pass of
its own, because it is read off the value forward. A module trained with
mixed-precision Adam holds $16P$ bytes of weights, gradients, master weights and
moments, against $2P$ when frozen, so resident model state falls from $32P$ to
$18P$ (44\%); at $P{=}4$B with 8-way sharding the critic's share is 7.0\,GB per
GPU.

\begin{figure}[t]
\centering
\includegraphics[width=0.84\textwidth]{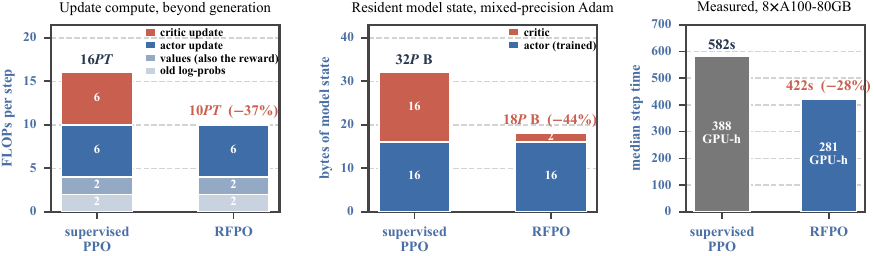}
\caption{\textbf{What freezing the critic removes.} Left and center are
accounting rather than measurement and hold independent of model scale and
hardware: $P$ is the parameter count shared by policy and critic, $T$ the tokens
in a batch, a forward pass costs ${\approx}2PT$ and a forward--backward
${\approx}6PT$, and a module trained with mixed-precision Adam holds $16P$ bytes
against $2P$ when frozen. The value pass survives both cuts because \car reads
the reward off it. Right is measured on matched batches (3.9M against 3.8M
tokens per step), medians over the run, with peak reserved memory 9.4\,GB lower
per GPU. With activation recomputation the saving is 40\%, not 37\%
(Appendix~\ref{app:cost}).}
\label{fig:cost}
\end{figure}

\paragraph{Measurement.}
Table~\ref{tab:timing} gives median per-step times over the first 300 steps of
supervised PPO and the first zero-label run, on matched batches (3.87M and 3.79M
tokens per step). Generation and every shared pass take the same time in both
arms; the whole difference is the critic update, 151\,s, which is close to the
actor update's 160\,s as the accounting predicts. Step time falls from 582 to
421\,s ($-28\%$), throughput rises from 828 to $1{,}122$ tokens per second
($+35\%$), and peak reserved memory falls by 9.4\,GB per GPU, the predicted 7.0\,GB
plus about 2.4\,GB of activations held by the critic's backward pass. A 300-step
run costs 388 GPU-hours under supervised PPO and 281 under \car, on one node of
eight A100-80GB GPUs. Under the single-update rule the old-log-probability pass
computes the same numbers as the actor's own forward, so a further 42\,s per step
could be saved; we left it in place to keep the two arms identical.

\begin{table}[h]
\centering
\footnotesize
\setlength{\tabcolsep}{5pt}
\caption{\textbf{Measured step time and memory.} Medians over steps 1--300;
validation and checkpointing are excluded.}
\label{tab:timing}
\begin{tabular}{lrr}
\toprule
\rowcolor{hdrbg}
Per-step component (median seconds) & PPO, 100\% labels & \car, 0\% labels \\
\midrule
Generation & 181 & 176 \\
Old log-probabilities (forward) & 43 & 42 \\
Values (forward; also the reward in \car) & 38 & 37 \\
Advantages and weight sync & 6 & 6 \\
Actor update & 160 & 156 \\
Critic update & 151 & --- \\
\midrule
Step & 582 & \textbf{421} \\
Throughput (tokens/s) & 828 & \textbf{1{,}122} \\
Peak memory, allocated / reserved (GB per GPU) & 22.0 / 32.4 & 18.9 / 23.0 \\
\bottomrule
\end{tabular}
\end{table}

\paragraph{One-off preparation.}
The savings apply to the RL stage. Obtaining the critic took 800 steps of
supervised PPO, 901 GPU-hours of training steps plus 38 of validation, which also
produced the initial policy that every arm starts from. Calibration is a single
scoring pass of the frozen critic over 512 rollouts.

\section{Run roster}
\label{app:roster}

Table~\ref{tab:roster} maps every run in the paper to its training log in the
supplementary material (\texttt{data/logs/}). Calibration files are in
\texttt{data/critic/baselines/}: the deployed length baseline and threshold are
\texttt{len\_baseline\_v2.json}, and the continuous arms use the
\texttt{len\_baseline\_v2\_finished*} family; the $4{,}096$-token run uses
\texttt{len\_baseline\_v2\_cap4096\_k1.7.json}, and the curves of
Figure~\ref{fig:cap4096} are in \texttt{data/derived/cap4096/}. The 74 runs of
Appendix~\ref{app:inner} are summarized in
\texttt{data/derived/stability/runs\_summary.csv}.

\begin{table}[!h]
\centering
\footnotesize
\setlength{\tabcolsep}{4pt}
\caption{\textbf{Run roster.} Every RL run starts from the initial policy and
frozen critic extracted from the phase-2 run, except the raw-score run, which
used an earlier actor--critic pair and an $8{,}192$-token cap.}
\label{tab:roster}
\begin{tabular}{lll}
\toprule
\rowcolor{hdrbg}
Run & Log & Used in \\
\midrule
SFT & \texttt{sft\_4n\_4581861} & App.~\ref{app:recipes} \\
Critic pretraining, phase 1 & \texttt{h061\_warmstart\_phase1\_ppo} & \S\ref{sec:identity}, App.~\ref{app:recipes}, \ref{app:critic} \\
Critic pretraining, phase 2 & \texttt{h065\_warmstart\_phase2\_ppo} & \S\ref{sec:identity}, App.~\ref{app:recipes}, \ref{app:critic} \\
\midrule
PPO, 100\% labels & \texttt{h066\_label100\_ppo} & \S\ref{sec:results}, App.~\ref{app:eval}--\ref{app:critic} \\
\car, 50\% labels & \texttt{h054\_label50\_car} & \S\ref{sec:results}, App.~\ref{app:eval}--\ref{app:critic} \\
\car, 0\% labels, run 1 & \texttt{h056\_label0\_seed2\_car} & \S\ref{sec:results}, App.~\ref{app:eval}--\ref{app:critic} \\
\car, 0\% labels, run 2 & \texttt{stage2\_my-host\_065} & \S\ref{sec:results}, App.~\ref{app:eval}, \ref{app:dynamics} \\
\car, 0\% labels, run 3 & \texttt{stage2\_my-host\_071} & \S\ref{sec:results}, App.~\ref{app:eval}--\ref{app:critic} \\
\midrule
Raw score & \texttt{stage2\_my-host\_044} & App.~\ref{app:ladder} \\
Threshold only, $\mathbf{1}[v>0.6]$ & \texttt{stage2\_my-host\_046} & App.~\ref{app:ladder} \\
Continuous, debiasing 0\% & \texttt{stage2\_my-host\_089} & \S\ref{sec:results}, App.~\ref{app:reward} \\
Continuous, debiasing 16\% & \texttt{stage2\_my-host\_085} & \S\ref{sec:results}, App.~\ref{app:reward} \\
Continuous, debiasing 34\% & \texttt{stage2\_my-host\_086} & \S\ref{sec:results}, App.~\ref{app:reward}, \ref{app:drift} \\
Continuous, debiasing 56\% & \texttt{stage2\_my-host\_083} & \S\ref{sec:results}, App.~\ref{app:reward}, \ref{app:drift} \\
Continuous, 56\%, group-relative & \texttt{stage2\_my-host\_088} & App.~\ref{app:grpo} \\
$1{,}024$-token cap & \texttt{stage2\_my-host\_048} & \S\ref{sec:results}, App.~\ref{app:budget} \\
\car, $4{,}096$-token cap & \texttt{stage2\_my-host\_098} & \S\ref{sec:results}, App.~\ref{app:budget} \\
PPO, $4{,}096$-token cap & \texttt{stage2\_my-host\_099} & \S\ref{sec:results}, App.~\ref{app:budget} \\
\bottomrule
\end{tabular}
\end{table}

\end{document}